\documentclass[journal]{IEEEtran}
\usepackage{subfigure}
\usepackage{amsmath,amsfonts}
\usepackage{algorithmic}
\usepackage{algorithm}
\usepackage{array}
\usepackage[caption=false,font=normalsize,labelfont=sf,textfont=sf]{subfig}
\usepackage{textcomp}
\usepackage{stfloats}
\usepackage{url}
\usepackage{verbatim}
\usepackage{threeparttable}
\usepackage{graphicx}
\usepackage{cite}
\usepackage[usenames, dvipsnames]{color}
\newtheorem{definition}{Definition}
\newtheorem{theorem}{Theorem}
\usepackage[misc]{ifsym}
\usepackage{enumitem}
\usepackage{bbding}
\usepackage{bm}
\usepackage{multirow}
\usepackage[switch]{lineno}

\usepackage{xcolor}


\begin{document}

\title{Computation and Parameter Efficient Multi-Modal Fusion Transformer for Cued Speech Recognition}

        % <-this % stops a space

\author{Lei Liu$^*$, Li Liu$^{*}$,~\IEEEmembership{Member,~IEEE}, Haizhou Li,~\IEEEmembership{Fellow,~IEEE}  % <-this % stops a space
\thanks{Lei Liu is with The Chinese University of Hong Kong, Shenzhen, Guangdong 518060, China, and also with Shenzhen Research Institute of Big Data, Shenzhen, Guangdong 518060, China. }
\thanks{Li Liu is with The Hong Kong University of Science and Technology (Guangzhou), Guangdong 511458, China (Corresponding author, email: avrillliu@hkust-gz.edu.cn).}
\thanks{Haizhou Li is with the Shenzhen Research Institute of Big Data, School of Data Science, Chinese University of Hong Kong, Shenzhen 518172, China, also with the University of Bremen, 28359 Bremen, Germany, also with the National University of Singapore, Singapore 119077.}
\thanks{$*$ indicates the equal contribution for the first two authors.}

}

\markboth{IEEE Transactions on Audio, Speech, and Language Processing}%
{Shell \MakeLowercase{\textit{et al.}}: A Sample Article Using IEEEtran.cls for IEEE Journals}

\IEEEpubid{0000--0000/00\$00.00~\copyright~2021 IEEE}
% Remember, if you use this you must call \IEEEpubidadjcol in the second column for its text to clear the IEEEpubid mark.

\maketitle
% \linenumbers
\begin{abstract}

Cued Speech (CS) is a pure visual coding method used by hearing-impaired people that combines lip reading with several specific hand shapes to make the spoken language visible. Automatic CS recognition (ACSR) seeks to transcribe visual cues of speech into text, which can help hearing-impaired people to communicate effectively. The visual information of CS contains lip reading and hand cueing, thus the fusion of them plays an important role in ACSR. However, most previous fusion methods struggle to capture the global dependency present in long sequence inputs of multi-modal CS data. As a result, these methods generally fail to learn the effective cross-modal relationships that contribute to the fusion. Recently, attention-based transformers have been a prevalent idea for capturing the global dependency over the long sequence in multi-modal fusion, but existing multi-modal fusion transformers suffer from both poor recognition accuracy and inefficient computation for the ACSR task. To address these problems, we develop a novel computation and parameter efficient multi-modal fusion transformer by proposing a novel Token-Importance-Aware Attention mechanism (TIAA), where a token utilization rate (TUR) is formulated to select the important tokens from the multi-modal streams. More precisely, TIAA firstly models the modality-specific fine-grained temporal dependencies over all tokens of each modality, and then learns the efficient cross-modal interaction for the modality-shared coarse-grained temporal dependencies over the important tokens of different modalities. Besides, a light-weight gated hidden projection is designed to control the feature flows of TIAA. The resulting model, named \textit{Eco}nomical \textit{Cued} Speech Fusion Transformer (\textit{EcoCued}), achieves state-of-the-art performance on all existing CS datasets (\textit{i.e.}, Mandarin Chinese, French, and British CS), compared with existing transformer-based fusion methods and ACSR fusion methods. Notably, our method dramatically reduces the computational complexity from $\mathcal{O}(T^2)$ to $\mathcal{O}(T)$. We will release the source code and data as open source.
\end{abstract}

\begin{IEEEkeywords}
Transformer, Cross-attention, Automatic Cued Speech Recognition, Computation and Parameter Efficient.
\end{IEEEkeywords}

\section{Introduction}
\begin{figure}[t]
\centering
\includegraphics[width=1\linewidth]{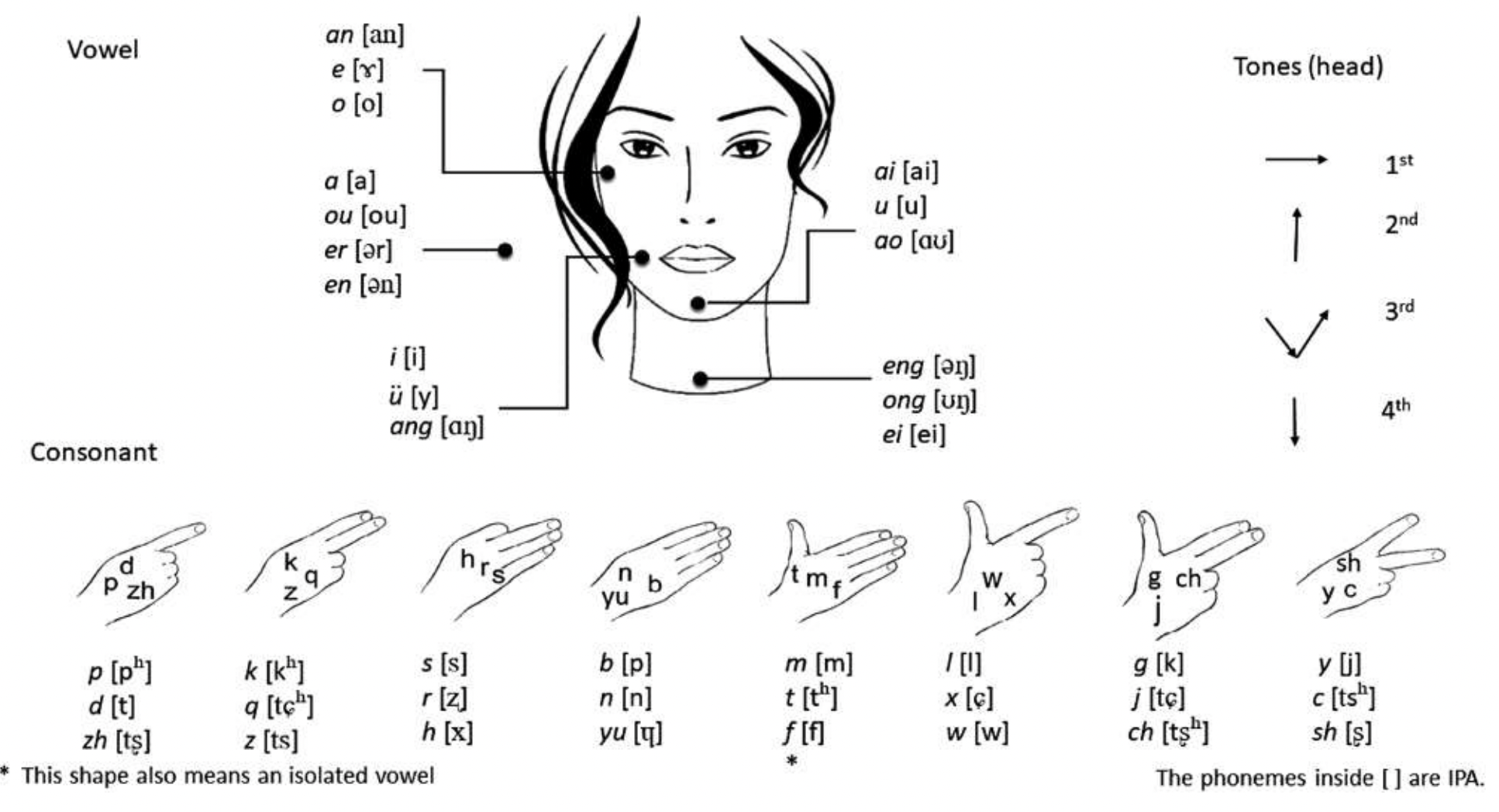}
\caption{The Mandarin Chinese CS system (image from \cite{liu2019pilot}). Combined with lip reading, five hand positions (mouth, chin, throat, side, cheek) are defined to encode Chinese vowels and eight hand shapes to encode Chinese consonants.}
\label{chinese}
\end{figure}

\IEEEPARstart{I}{n} order to address the insufficient information of lip reading and enhance the reading skills of hearing-impaired children, in 1967, Cornett \cite{cornett1967cued} invented the first Cued Speech (CS) system for American English to use hand codings to complement lip reading in phonetic level, making the spoken language visible. American CS employs four hand positions and eight hand shapes based on two main criteria: minimal effort for spoken speech encoding and maximum visual contrast for good speech perception. Later, CS has been adapted to more than 65 spoken languages. In 2019, Liu \textit{et al.} \cite{liu2019pilot} proposed the first Mandarin Chinese CS system (see Figure \ref{chinese}), where five hand positions (mouth, chin, throat, side, cheek) were defined to encode all Chinese vowel groups and eight hand shapes to encode Chinese consonant groups.

With the advent of deep learning, Automatic Cued Speech Recognition (ACSR) \cite{liu2018automatic,liu2019automatic,zhang2023cuing} attracted increasing interests as it can potentially aid the hearing-impaired in daily communication. ACSR aims to transcribe multi-modal inputs (\textit{i.e.}, lip and hand movements) in a CS into text, where an appropriate cross-modal fusion strategy is essential to handle the complementary relationships from the multi-modal inputs.

% \textcolor{blue}{because lip reading and hand shape/position movements in ACSR follow different coding rules \cite{attina2004pilot} and they are weakly synchronous \cite{liu2020re,liu2018automatic} in temporal, \textit{i.e.}, lip-hand movements are synchronous at the utterance level and asynchronous at the word level.}

\IEEEpubidadjcol Existing studies for the multi-modal fusion in ACSR mainly focus on extracting and concatenating discriminative multi-modal features. For instance, \cite{heracleous2012continuous,liu2018visual,wang2021attention} marked the region of interest (ROI) of lip and hand to extract the visual features and directly concatenated these features for cross-modal fusion. \cite{liu2020re} proposed a re-synchronization procedure for multi-modal alignment, which needs to statistically pre-define the hand preceding time of the CS dataset. \cite{wang2021cross} exploited knowledge distillation to extract effective features from teacher knowledge of the speech data. However, these methods did not consider the global dependency over the long sequence inputs of the CS data. Therefore, these methods generally failed to effectively characterize the multi-modal inputs for the cross-modal fusion.
\newpage
% MSHMM \cite{liu2018visual} manually assigned importance weights for different modalities to achieve adaptive fusion of lips and hands. 
% \textcolor{blue}{However, these methods did not consider the global dependency over the long sequence inputs of the CS data. Besides, they generally ignored the weakly synchronous issue or required pre-defined statistical information for mutual alignment in the ACSR task. Therefore, all these methods failed in effectively characterizing the multi-modal inputs for cross-model fusion under the weakly synchronous issue.} 
%%%% CS中的多模态融合问题---》由于CS多模态异步性，导致这个融合是有挑战的---》之前的方法要么先对齐再简单的串联，或者用了交叉注意力，但是这些方法有什么缺点。
Recently, transformers have been proven to achieve good performance for multi-modal tasks, since they can utilize cross-attention mechanism to capture the latent cross-modal similarity \cite{wei2020multi} with global dependency \cite{vaswani2017attention,qian2022audio}. To this end, \cite{liu2023cross} proposed a cross-modal mutual learning method based on the transformer to handle the multi-modal fusion in ACSR. However, this method is computationally and parameter costly.

\begin{figure*}[t]
\centering
\subfigure[Previous Transformers for Multi-modal Fusion]{\includegraphics[width=0.409\linewidth]{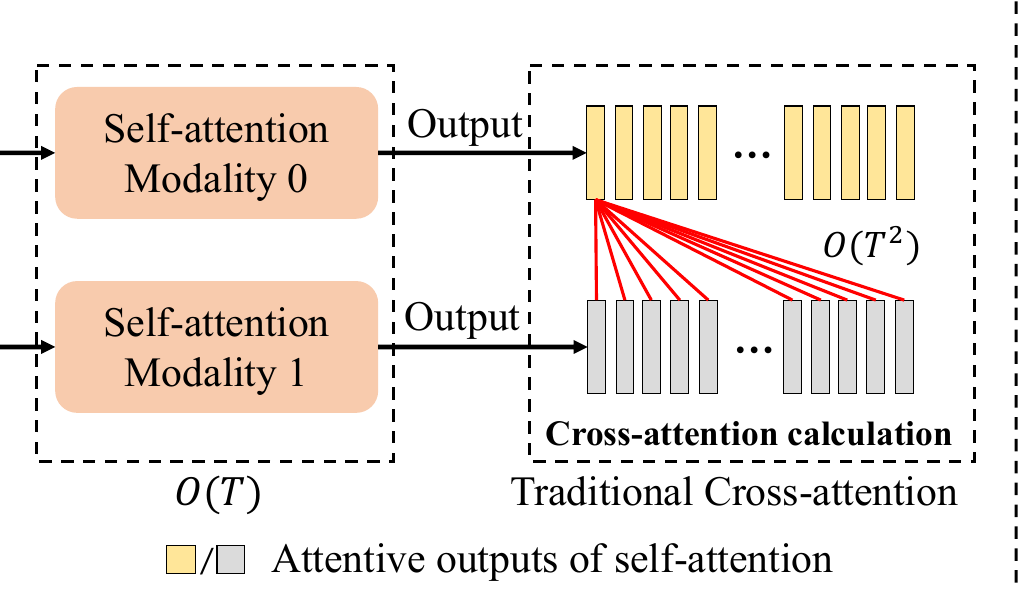}}\hspace{4pt}
\subfigure[Our Method for Multi-modal Fusion]{\includegraphics[width=0.565\linewidth]{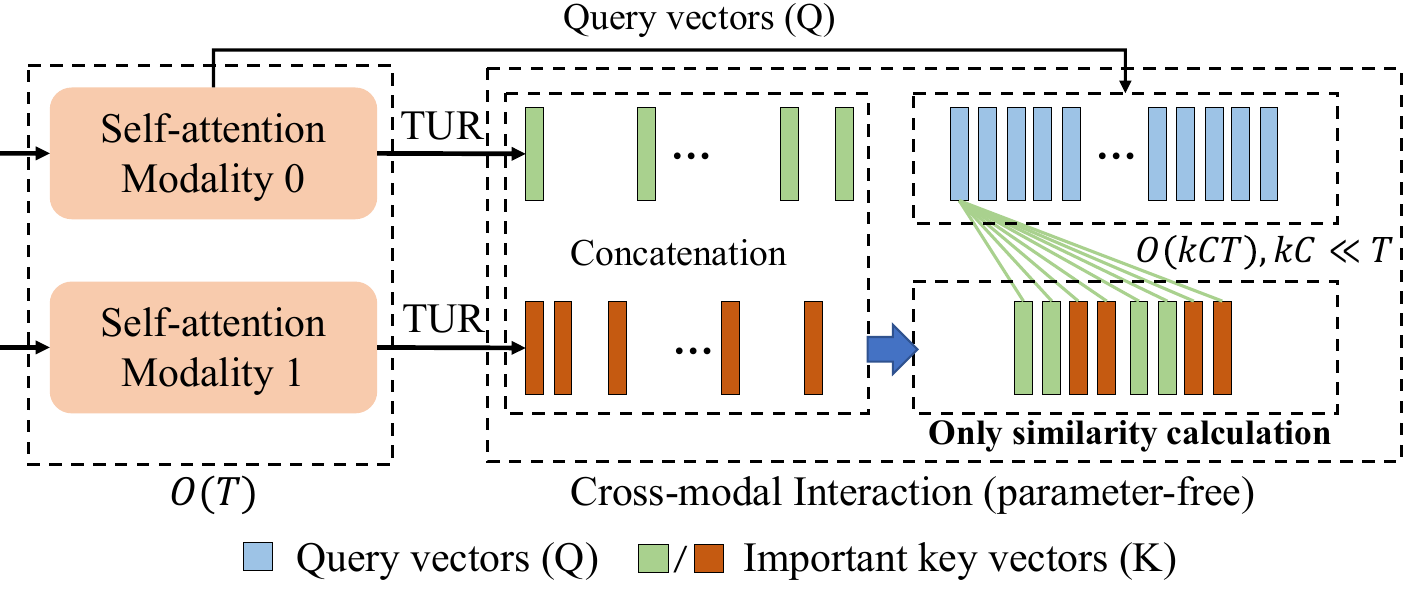}}
\caption{Multi-modal fusion comparison between previous transformers (left) and the proposed method (right). $T$ is the input sequence length. $C$ is the chunk number for segmenting the input sequence. $k$ is the number of selected important tokens in each chunk. \textbf{(a)} Previous transformers would introduce extra computation and parameters for cross-modal interaction, requiring quadratic complexity (red links) and projection layers. \textbf{(b)} Our method utilizes a parameter-free cross-modal interaction with linear computation complexity (green links). Here $k=2$ is the simplest case for the visualisation purpose.}
\label{com_fusion}
\end{figure*}

In the previous literature, various methods have been explored to decrease the model complexity (\textit{i.e.}, computation complexity and parameter amount) of the transformer \cite{katharopoulos2020transformers,mehtadelight,wu2022flowformer}. In fact, the efficiency bottlenecks of the transformer mainly come from the quadratic complexity of the self-attention \cite{peng2020random} and the large parameters of the feed-forward networks \cite{mehtadelight}. More precisely, self-attention requires each token to attend to all other tokens via the dot product operation \cite{vaswani2017attention}, resulting in quadratic complexity over the input length. Previous efficient techniques generally rely on the following essential properties \cite{tay2022efficient} to decrease the complexity of the self-attention: (i) Softmax-based score elements of the attention matrix are non-negative \cite{katharopoulos2020transformers}. Thus the softmax operation can be approximated in a similar but more efficient way, such as kernel function \cite{qin2021cosformer}, positive random features \cite{choromanski2020rethinking}, and random Fourier features \cite{peng2020random}. (ii) The softmax-based attention matrix is low-rank \cite{wang2020linformer}. The particular solution is to introduce the sparsity property into attention matrices \cite{child2019generating,kitaev2019reformer,zaheer2020big,wang2020linformer}. Besides, the feed-forward network is exploited to improve the expressiveness of transformers, but it introduces more parameters via many stacked fully-connected layers. To be more light-weight, some efficient architectures with fewer parameters are utilized to replace the feed-forward network in the transformer, such as convolutions \cite{wupay}, gated linear units \cite{dauphin2017language}, and multi-branch feature extractors \cite{wulite}.

\begin{figure}[t]
\begin{center}
\centerline{\includegraphics[width=0.8\columnwidth]{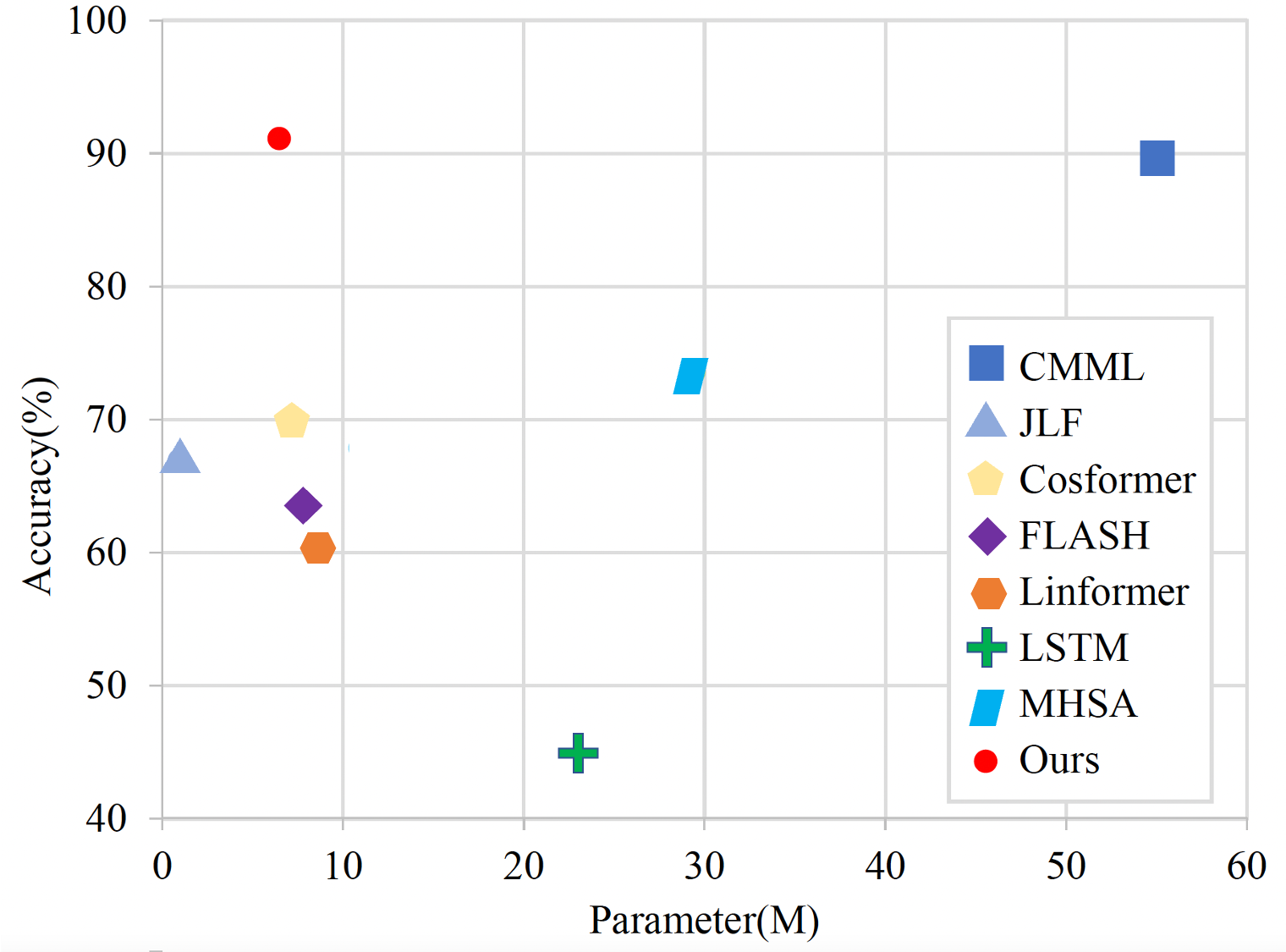}}
\caption{Phoneme-level recognition accuracy on Chinese CS dataset with respect to parameters. Comparison with ACSR methods: LSTM \cite{papadimitriou2021fully}, JLF \cite{wang2021cross}, and CMML \cite{liu2023cross}; Comparison with Transformer models: vanilla Multi-Head Self-Attention (MHSA) \cite{vaswani2017attention}, FLASH \cite{hua2022transformer}, Linformer \cite{wang2020linformer}, Performer \cite{choromanski2020rethinking}, and Cosformer \cite{qin2021cosformer}. RegNet \cite{Radosavovic_2020_CVPR} is the front-end backbone for all methods.}
\label{acc_para}
\end{center}
\end{figure}

Although prior studies of efficient transformers \cite{hua2022transformer,tay2022efficient} have achieved the self-attention with linear computational complexity using light-weight architectures (\textit{e.g.}, FLASH in \cite{hua2022transformer}), few works focused on the multi-modal fusion. When directly applying these methods to the multi-modal fusion for ACSR, it remains some significant challenges and may introduce extra computation and parameter. More specifically, these works often only capture long-time dependencies for a single-modality sequence using an individual attention flow, and then a cross-modal interaction, which is conducted by feature concatenation \cite{liu2018automatic,liu2020re} or the cross-attention mechanism (\textit{e.g.}, the visual-linguistic alignment module in \cite{liu2023cross}). We present Figure \ref{com_fusion} to illustrate the multi-modal fusion comparison between previous transformers and our method. Due to the lack of effective yet efficient cross-modal fusion for enhancing spatial-temporal relations of different modalities, previous attention-based fusion approaches often suffer from significant performance drops for the ACSR task. 
As shown in Figure \ref{acc_para}, it can be seen that the previous attention-based fusion methods (with more parameters) perform even worse than our method (with fewer parameters) in phoneme-level ACSR recognition accuracy. Therefore, it is necessary to develop an efficient and effective transformer-based fusion method with low model complexity for ACSR. 
%\textbf{Therefore, it is necessary to develop an efficient and effective transformer-based fusion method with low model complexity for ACSR.}}

% our method can achieve a parameter-free multi-modal fusion with lower computation complexity compared with .
% At the same time, when applying these existing approaches to the ACSR task, they would inevitably introduce additional computation costs or model parameters for the multi-modal fusion, \textit{i.e.}, expanded feature dimension or the weights in cross-attention modules, making it difficult to apply these models to ACSR in real-world computation resources limited scenarios. 

% due to quadratic complexity with respect to the input length, 
% When dealing with such a multi-modal task, most of these works utilize an individual attention flow to capture long-time dependencies for a single-modality sequence, and then they simply concatenate multi-modal features \cite{liu2018automatic,liu2020re} or exploit the cross-attention mechanism \cite{shukor2022transformer} as the main multi-modal fusion strategy. Due to the lack of effective cross-modal fusions for enhancing spatial-temporal relations of different modalities in CS, these approaches often suffer from significant performance drops for the ACSR task, \textit{e.g.}, worse phoneme-level recognition accuracy compared with our proposed method as shown in Figure \ref{acc_para}.

In this work, we propose a novel efficient attention-based transformer architecture for the multi-modal fusion in automatic CS recognition (ACSR) called \textit{Eco}nomical \textit{Cued} Speech Fusion Transformer (\textit{EcoCued}). The whole framework is illustrated in Figure \ref{frame}. Motivated by the low-rank property of the self-attention, a novel Token-Importance-Aware Attention mechanism (TIAA) is proposed to model the long-time dependencies over the multi-modal CS inputs, where a token\footnote{A input CS video can be mapped into a frame-wise feature space by a front-end, where the feature of one frame is called one token in this work.} utilization rate (TUR) is designed to select important tokens from each modality. Concretely, TIAA decomposes the full self-attention into \textbf{modality-specific} and \textbf{modality-shared} components to capture local and global temporal dependencies from different modalities. Besides, TIAA achieves an effective multi-modal fusion for the modality-shared component by fusing the important tokens of different modalities. Based on such an attention mechanism, a Convolution-based Aggregation (ConAgg) module is presented to achieve spatial interaction for modality-specific and modality-shared components. Finally, instead of the feed-forward network, a light-weight gated hidden projection is designed to control the feature flow through the TIAA module, allowing the model to focus on the most important features for the ACSR task. \textbf{In summary, the key contributions of this work are the following threefold:}

% TIAA decomposes the full self-attention into \textbf{modality-specific} and \textbf{modality-shared} components. Modality-specific component models the local fine-grained temporal dependencies of the tokens within each chunk for each modality, while the modality-shared component captures the global coarse-grained temporal dependencies of important tokens for different modalities. 
\begin{itemize}
    \item 
    To address the efficiency issue of ACSR, we propose a novel computation and parameter efficient multi-modal fusion transformer called \textit{EcoCued}, which can capture both long-time dependencies by the proposed novel TIAA and spatial relations by the ConAgg. 
    \item 
    We propose a token utilization rate (TUR) to select the important tokens from each modality. TUR-based TIAA can decompose the full self-attention into modality-specific and modality-shared components for unimodal fine-grained dependency and cross-modal coarse-grained dependency, respectively.
    %On this basis, we achieve an efficient and effective cross-modal interaction by fusing the important tokens of different modalities. 
    \item
    Compared with existing efficient attention-based fusion methods and previous fusion methods in ACSR, the proposed EcoCued can achieve SOTA performance on all existing CS datasets (\textit{i.e.}, Mandarin Chinese, French, and British English CS datasets). Notably, our method reduces the computational complexity of the self-attention from $\bm{\mathcal{O}(T^2)}$ to $\bm{\mathcal{O}(T)}$ with a light-weight transformer-based architecture. Compared with the previous SOTA method in ACSR \cite{liu2023cross}, our method can significantly reduces the parameter number of the model from \textbf{54.9M} to \textbf{6.6M}.
    % Compared with existing efficient attention methods and previous ACSR methods, the proposed EcoCued achieves SOTA performance on three CS benchmarks (\textit{i.e.}, Chinese, French, and British English CS). Importantly, our method reduces the computational complexity from $\mathcal{O}(T^2)$ to $\mathcal{O}(T)$ in both time and space levels.
\end{itemize}

\section{Related Work}
In this section, we first provide an overview of the relevant works for multi-modal fusion in ACSR. Then, we discuss the recent progress for the efficient transformer.

\subsection{Multi-modal Fusion in ACSR}
Recently, multi-modal learning is demonstrated to be effective for speech processing and natural language processing (NLP) tasks, such as multi-modal speech emotion recognition \cite{siriwardhana2020jointly,priyasad2020attention}, spoken language understanding \cite{you2020towards,you2021knowledge,you2022end,you2020contextualized,chen2021self}. For instance, \cite{you2021self} proposed a temporal-alignment attention to align the speech-text feature clues for the spoken question answer tasks. \cite{you2021mrd} proposed a multi-modal residual knowledge distillation method to adaptively leverage audio-text features. By considering global dependency for multi-modal interactions, these methods could obtain superior performance for their corresponding tasks. Motivated by this, we mainly focus on efficiently capturing global dependency to enhance the contextual understanding in continuous CS videos for the ACSR task.

Multi-modal fusion is an important step in automatic CS recognition to capture complementary relationships between lip and hand movements. Early studies of ACSR tended to directly concatenate the features of multi-modal inputs as the dominant fusion paradigm. For example, \cite{burger2005cued,stillittano2013lip} used different colors to mark lip and hand regions for further feature extraction and fusion, as well as the coordinates of the marks on the finger. The regions of interest (ROIs) were segmented to extract the ROI-based features of lips and hands, which exploited a pre-defined threshold to track the marks of cuers \cite{heracleous2010cued,heracleous2012continuous}. Recent works \cite{liu2020re} gradually get rid of such artifices on the lips and hands. For instance, MSHMM \cite{liu2018visual} merged different features by giving weights manually for different CS modalities, and \cite{wang2021cross} adopted knowledge distillation for better unimodal representations. In order to learn a better fusion strategy, \cite{liu2019novel,liu2020re} proposed shifting the hand movement sequence with a statistically computed value to align semantically with lip movements before concatenating them for fusion. However, these methods ignored the global dependency present in the long sequence inputs of CS data, resulting in limited interactions of multi-modal inputs for cross-modal relation capturing. In order to address the above-mentioned global dependency problem, Liu \textit{et al.} \cite{liu2023cross} introduced a transformer-based approach to learn modality-invariant shared linguistic representations that guide the semantic alignment of multi-modal data streams at the phonetic level. However, this method encounters challenges related to huge computational complexity and parameter requirements.

% while previous methods generally assumed lip-hand movements are sequentially synchronous by default. To address the asynchronous issue, \cite{liu2020re} proposed a re-synchronization procedure to align visual features based on the prior of hand preceding time, which is dependent on statistical information of CS cuers and datasets. \cite{liu2023cross} proposed to learn modality-invariant shared linguistic representations to guide the alignments for multi-modal data streams. However, existing studies either focused on small-scale datasets or ignored the efficient issue of the recognition model, which influences the real-world application of the ACSR task.

\begin{figure*}[!t]
	\centering
	\includegraphics[width=0.9\linewidth]{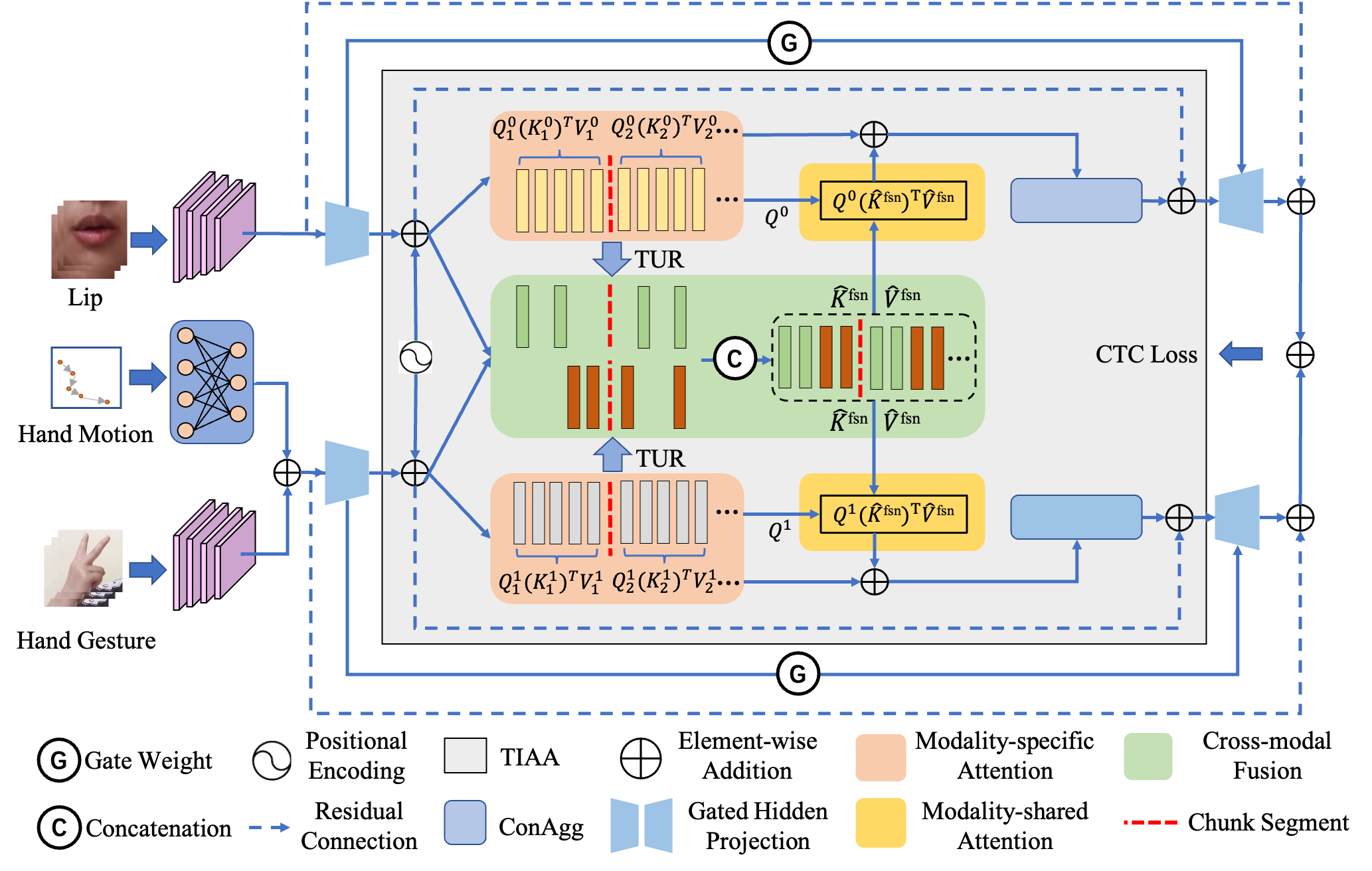}
  \caption{The illustration of the EcoCued approach. At first, pre-trained extraction models (dlib \cite{dlib} and mediapipe \cite{mediapipe}) are used to capture the ROIs of lip and hand from the videos. Then a shared frond-end \cite{Radosavovic_2020_CVPR} is utilized to extract frame-wise features for lip motions and hand shapes, and a linear layer is to extract features of hand positions. To reduce the complexity of self-attention, TUR is presented to select important tokens from each modality. The proposed TIAA mechanism first calculates the modality-specific attention to capture the local fine-grained dependencies within each chunk of the sequence for each modality. Then, TIAA fuses the important tokens of different modalities and calculates the modality-shared coarse-grained dependencies over the fused tokens. Finally, a convolution aggregation (\textit{i.e.}, ConAgg) module is used to aggregate the modality-specific and modality-shared attention flows along with the spatial dimension. Besides, gate hidden projection is presented to control the information flow from input to output projections for TIAA.} 
  % 下面这个是小修之后的版本
  % \caption{The illustration of the EcoCued approach. At first, pre-trained extraction models (dlib \cite{dlib} and mediapipe \cite{mediapipe}) are used to capture the ROIs of lip and hand from the videos. Then a shared frond-end \cite{Radosavovic_2020_CVPR} is utilized to extract frame-wise features for lip motions and hand shapes, and a linear layer is to extract features of hand positions. To reduce the complexity of self-attention, TUR is presented to select important tokens from each modality. The proposed TIAA mechanism first calculates the modality-specific attention to capture the local fine-grained dependencies within each chunk of the sequence for each modality. Then, TIAA fuses the important tokens of different modalities and calculates the modality-shared coarse-grained dependencies over the fused tokens. Finally, a convolution aggregation (\textit{i.e.}, ConAgg) module is used to aggregate the modality-specific and modality-shared attention flows along with the spatial dimension. Besides, gate hidden projection is presented to control the information flow from input to output projections for TIAA.}
\label{frame}
\end{figure*}

% and further revealed two critical challenges of ACSR: (1) Cross-modal interaction, where lip and hand sequences are complementary to each other as two distinct modalities. (2) Modality asynchronism caused by hand preceding phenomenon \cite{liu2020re}.  

% \subsection{Computation-Efficient Self-attention}
\subsection{Computation-Efficient Transformer}
There are many prior studies on addressing the efficiency bottleneck of the transformer \cite{tay2022efficient}. Most approaches work towards decreasing the quadratic complexity of self-attention, and few studies focus on multi-modal fusion \cite{xu2023multimodal}. In this part, we will review two common techniques for the efficient self-attention including sparsity and similarity approximation.

\noindent\textbf{Sparse Attention.} This technique improves the efficiency of self-attention by computing a spare attention matrix, \textit{i.e.}, each token only attends partial tokens instead of all tokens. For instance, in the Sparse Transformer \cite{child2019generating}, the context attention matrix is computed between each token and its neighbor tokens, reducing the complexity from $\mathcal{O}(T^2)$ to $\mathcal{O}(T\sqrt{T})$. Furthermore, the tokens can be divided into multiple blocks to formulate blockwise self-attention \cite{qiu2019blockwise}, where quadratic complexity only happens for the selected blocks. \cite{kitaev2019reformer} proposed Reformer to reduce the complexity from $\mathcal{O}(T^2)$ to $\mathcal{O}(T\log T)$ using locality-sensitive hashing (LSH) for dot-product attention. \cite{vyas2020fast} proposed a clustered attention to group queries into different clusters and only computed attention for the centroids with linear complexity. \cite{zaheer2020big} further improved the sparse attention using the global tokens to achieve more effective information aggregation. However, these techniques suffer from significant performance degradation due to sacrificing information utilization with limited speed-up \cite{wang2020linformer}.

\noindent\textbf{Similarity Approximation.} This technique computes the attention matrix via the inner product between the non-linear projections (\textit{e.g.}, kernel functions) of queries and keys. For example, linear Transformer \cite{katharopoulos2020transformers} utilized the exponential linear unit as the non-linear projection. To approximate the softmax operator, Performer \cite{choromanski2020rethinking} considered positive random features and \cite{peng2020random} exploited the random Fourier features \cite{rahimi2007random} to compute the attention matrix, respectively. However, these approaches rely on specific kernels with approximate errors. Meanwhile, to avoid computing the full attention matrix, Nyström matrix decomposition \cite{williams2000using} is utilized in SOFT \cite{lu2021soft} and YOSO \cite{zeng2021you}. The cosine function is used in cosFormer \cite{qin2021cosformer} while generally introducing more calculation iterations or sacrificing the generality \cite{wu2022flowformer}.

Unlike prior methods, our method decomposes the full attention into modality-specific and modality-shared components, which capture fine-grained and coarse-grained dependencies for multi-modal inputs, respectively. Then a convolution aggregation module is performed to enhance the spatial interaction of the multi-modal contextual information. Based on this, we propose a flexible multi-modal fusion strategy by fusing the importance tokens of different modalities, which explicitly enjoys both linear complexity and effective cross-modal information interaction.

\section{Preliminaries}
\subsection{Problem Formulation} 
A CS dataset consists of $N$ quadruples of the lip, hand shape, hand position, and sentence-level label sequences, denoted by $\mathcal{D}=\{(X_i^l, X_i^g, X_i^p, Y_i)\}^N_{i=1}$, where lip and hand are complementary to each other as different modalities. The target is to train a model mapping multi-modal data streams $(X^l, X^g, X^p)$ into the corresponding linguistic sentence $Y$. Given the input sequences $(X^l, X^g, X^p)$ of length $T$, a CNN-based front-end is firstly employed to extract frame-wise representations $F_l, F_g, F_p \in \mathbb{R}^{T\times d_m}$, where $d_m$ is the representation dimension. Then element-wise addition operation $\oplus$ is conducted to fuse features of hand shape and position via $F_h=F_g \oplus F_p$. In simplification, $m\in \{0, 1\}$ denotes lip ($0$) and hand ($1$) modalities in the following section, respectively. For the rest of this paper, we will omit the subscript of $m$ except for the section \ref{multimodal}. Our work focuses on achieving an efficient multi-modal transformer with an effective cross-modal fusion strategy for ACSR, which captures both long-time temporal dependencies and spatial relations over the sequential representations of lip and hand modalities.

\subsection{Motivation}
In this section, we will review the Multi-Head Self-Attention (MHSA) \cite{vaswani2017attention} and experimentally demonstrate the low-rank property of MHSA for the ACSR task, motivating us to select the important tokens to improve the model efficiency.

% The traditional transformer architecture, \textit{i.e.}, stacked attention blocks (each block contains a MHSA layer and a feed-forward network), requires a quadratic complexity and a large number of parameters to model the long-time dependency \cite{vaswani2017attention}. Such an efficient issue is more aggravated when handling the multiple modalities by different attention branches, \textit{e.g.}, ACSR involving lip and hand modalities. Consequently, the transformer easily leads to the over-fitting issue due to the higher model complexity, especially when trained on a small dataset, \textit{e.g.}, the ACSR task with the relatively limited CS data. Therefore, it is important to develop an efficient yet effective transformer for multi-modal fusion for ACSR. Here, we simply review the MHSA as follows.

% \textcolor{red}{In particular, lip and hand movements in CS are asynchronous for the ACSR task.}
% (\textit{i.e.}, different heads)
\textbf{Multi-Head Self-Attention.}
Let's recall that the primary goal of transformers is to jointly aggregate tokens at different positions from multiple attention heads, where the MHSA operation is defined as:
\begin{equation}
\operatorname{MHSA}(Q, K, V)=\operatorname{concat}\left(\operatorname{HD}_1, \ldots, \operatorname{HD}_h\right) W^o \text {, }
\end{equation}
where $h$ is the head number and $Q, K, V \in \mathbb{R}^{T\times d}$ are input embedding matrices. $T$ is the sequence length and $d$ is the embedding dimension. $W^o\in \mathbb{R}^{d \times d}$ is the linear projection weight of the output layer. The self-attention operation is conducted within each subspace as follows:
\begin{equation}
\operatorname{HD}_i=\underbrace{\operatorname{softmax}\left[\frac{Q W_i^q\left(K W_i^k\right)^T}{\sqrt{d_{qk}}}\right]}_S V W_i^v,
\label{mhsa}
\end{equation}
where $W_i^q, W_i^k \in \mathbb{R}^{d \times d_{qk}}, W_i^v \in \mathbb{R}^{d \times d_v}$ are the linear projections for the subspace $\operatorname{HD}_i$ with the hidden dimensions as $d_{qk}$ and $d_{v}$. Self-attention calculates the scaled dot product between every query and key, which refers to a score matrix $S \in \mathbb{R}^{T \times T}$ with softmax-based normalized raws. 

As indicated above, the quadratic complexity of the self-attention arises from the sequence length (\textit{i.e.}, the number of tokens in a self-attention layer). Thus, to achieve an efficient transformer, a capable solution is to model the self-attention only over the important tokens in the sequence. Moreover, the cross-modal interaction also benefits from the fusion of the important tokens. In the following, we will exhibit the low-rank property of the self-attention for the ACSR task, indicating that the tokens corresponding to the largest singular values are important to recover full attention.

\begin{figure}[!t]
\begin{center}
\centerline{\includegraphics[width=0.7\columnwidth]{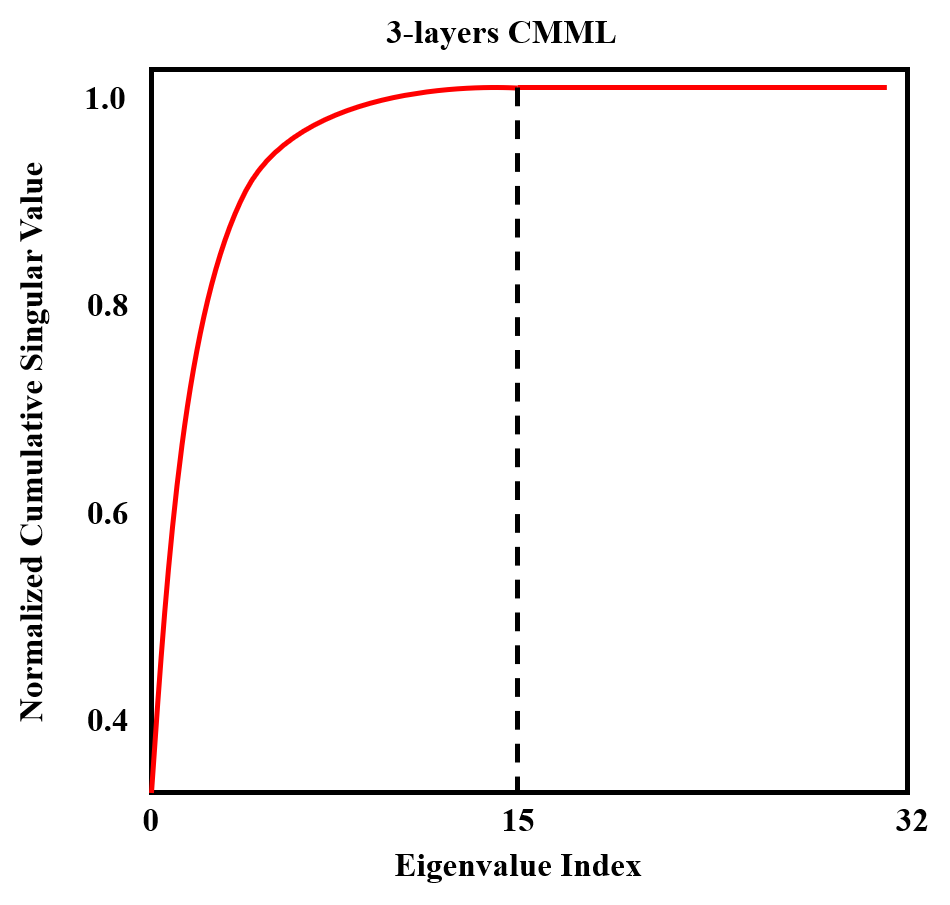}}
    \caption{Spectrum analysis of the self-attention matrix in the transformer \cite{liu2023cross} with top-128 largest eigenvalues. We can see the original MHSA formulation obtains a low-rank attention matrix for the ACSR task, which motivates us to focus on the most important tokens in the CS sequences.}
\label{singular}
\end{center}
\end{figure}

\textbf{Low-Rank Property.} In this part, we provide a spectrum analysis of the attention matrix $S$ for ACSR on the Chinese CS dataset, \textit{i.e.}, we apply singular value decomposition (SVD) for the attention matrix and plot the normalized cumulative singular value averaged over $1k$ sentences. As shown in Figure \ref{singular}, the spectrum curve exhibits a clear long-tail distribution, which indicates that only a few largest singular values can recover a large portion of information of the matrix $S$. \cite{wang2020linformer} provides the following theoretical results for the above spectrum analysis.

\begin{theorem} For any $Q, K, V \in \mathbb{R}^{T \times d}$ and $W_i^q, W_i^k, W_i^v \in \mathbb{R}^{d \times d}$, for any column vector $w \in \mathbb{R}^T$ of matrix $V W_i^v$, there exists a low-rank matrix $\tilde{S} \in \mathbb{R}^{T \times T}$ satisfying:
\begin{equation}
    \operatorname{Pr}\left(\left\|\tilde{S} w^T-S w^T\right\|<\epsilon\left\|S w^T\right\|\right)>1-o(1),
\end{equation}
where $\operatorname{rank}(\tilde{S})=\Theta(\log (T))$.
\label{theorem1}
\end{theorem}

According to Figure \ref{singular} and Theorem \ref{theorem1}, the self-attention formulation obtains a low-rank attention score matrix for the ACSR task. Therefore, it is feasible to focus on the important tokens in the sequence to reduce the complexity of the transformer. Motivated by this, we propose an EcoCued method with a novel TIAA for the ACSR task, which avoids performing an SVD decomposition in each attention matrix with additional complexity. Besides, the cross-modal interaction can be conducted efficiently by fusing important tokens.

\section{The Proposed Method}
In this section, we will first introduce the proposed EcoCued framework. Then, the TIAA mechanism will be described in detail, including modality-specific, modality-shared components, the defined TUR, and the cross-modal fusion. Then the ConAgg module is used to integrate modality-specific and modality-shared information via the spatial interaction along the spatial dimension of the features. The final one is for gated hidden projection to control the information flow of TIAA.

% In this section, we will first introduce the proposed EcoCued framework. Then, the TIAA mechanism will be described in detail, including modality-specific, modality-shared components, the defined TUR, and the cross-modal fusion. Then the following is about the ConAgg module to enhance the spatial relations for modality-specific and modality-shared components. The final one is for gated hidden projection to control the information flow of TIAA.

\begin{figure}[!t]
\centering
\includegraphics[width=0.7\linewidth]{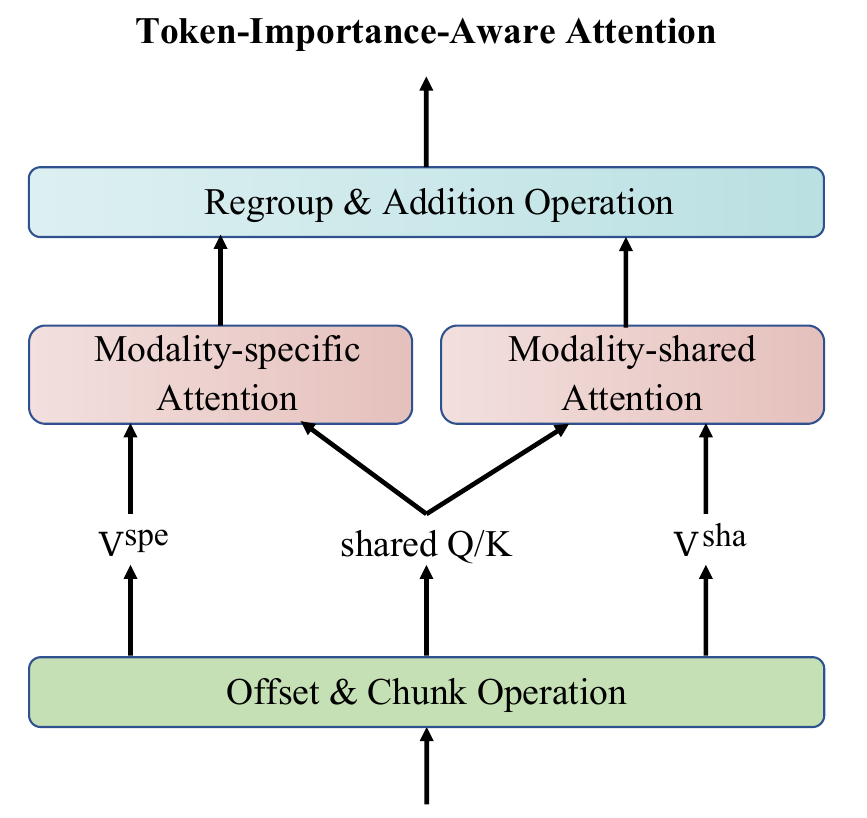}
\caption{The illustration of the TIAA mechanism. TIAA is composed of modality-specific and modality-shared attentions, which utilize shared query-key vectors (\textit{i.e.}, $Q/K$) with different value vectors (\textit{i.e.}, $V^\text{spe}$ for modality-specific attention and $V^\text{sha}$ for modality-shared attention), which can maintain the information gains from the different sparse patterns in TIAA.}
\label{TIAA}
\end{figure}

\textbf{EcoCued Framework.} The whole framework is illustrated in Figure \ref{frame}. For each modality, given the input sequence $F \in \mathbb{R}^{T\times d_m}$, the hidden embedding sequence $F_{u} \in \mathbb{R}^{T\times d}$ is firstly obtained by a gated hidden projection (introduced in Section \ref{GHP}). Then, as shown in Figure \ref{TIAA}, TIAA is used to decompose the full attention into modality-specific and modality-shared attentions for each modality:
\begin{equation}
\operatorname{HD}=A^{\text{spe}} V^{\text{spe}} + A^{\text{sha}} V^{\text{sha}},
\end{equation}
where the subscript for $\operatorname{HD}$ is omitted since our model only has one head space. $A^{\text{spe}}$ and $A^{\text{sha}}$ are attention score matrices for modality-specific and modality-shared branches with linear computational complexity, respectively. Note that all trainable parameters are shared for different modalities. Importantly, in the modality-shared branch, TIAA fuses the important tokens of different modalities, which are selected in each modality by the proposed TUR.

\subsection{Token-Importance-Aware Attention Mechanism}
% TIAA mechanism combines the benefits from both modality-specific fine-grained and modality-shared coarse-grained attentions, which can be complementary to each other by sharing the same architecture but focusing on temporal relationships of different time ranges and different modalities. 

TIAA mechanism benefits from the complementary roles of modality-specific and modality-shared attentions by sharing the same architecture for different modalities.

% The decomposition pattern of TIAA is illustrated in Figure \ref{idea}.
% In detail, the hidden embedding $F_{u}$ is separated into non-overlapping chunks of size $C$, \textit{i.e.}, $[T] \xrightarrow{} [T/C \times C]$.
\textbf{Chunk Operation.} Before TIAA calculation, a chunk operation \cite{hua2022transformer} is utilized to separate the sequence into different parts, which is parameter-free to decrease the computation complexity. In detail, the hidden embedding $F_{u}$ with length $T$ is separated into non-overlapping $(T\times C)$ chunks, where each chunk contains $C$ tokens. To avoid additional parameters for the projections of query, key, and value vectors, we adopt the per-dimension scaler and offset operations \cite{hua2022transformer} to accordingly produce $Q_{c}, K_{c}, V_{c}^{\text{spe}}, V_{c}^{\text{sha}}$ for $c$-th chunk. In particular, we can control the chunk size for the trade-off between performance and efficiency, referring to the Section \ref{ablation}. 

% \begin{figure}[!t]
% \begin{center}
% \centerline{\includegraphics[width=1\columnwidth]{idea.pdf}}
% \caption{The decomposition scheme of TIAA. We illustrate how the full attention map can be decomposed into modality-specific and modality-shared components, which are sparse matrices with different sparse patterns.}
% \label{idea}
% \end{center}
% \end{figure}

\textbf{Modality-specific Attention.} Since lip and hand in the ACSR task exhibit different visual cues (\textit{e.g.}, appearance, shape, and motion) to represent the same CS phoneme, modality-specific attention is independently applied to each modality to model their own fine-grained dependencies. Within $c$-th chunk for one modality, the modality-specific dependency is formulated as:
\begin{equation}
F_c^{\text{spe}}= A_c^{\text{spe}} V_c^{\text{spe}} = \psi\left(Q_c K_c^{T}\right) V_c^{\text{spe}},
\end{equation}
where $\psi$ is a regular activation function to replace the softmax operator and $A_c^{\text{spe}}$ is the modality-specific attention matrix for $c$-th chunk. This simplification \cite{hua2022transformer} is feasible in the case of using a gating mechanism (introduced in the section \ref{GHP}). Then the final attentive result $F^\text{spe}$ is obtained by re-grouping different local chunks:
\begin{equation}
F^\text{spe} = \operatorname{concat}(F_0^{\text{spe}}, F_1^{\text{spe}}, \cdots, F_{n-1}^{\text{spe}}),
\end{equation}
which concatenates the attentive results of each chunk and $n=T/C$. Note that modality-specific attention spends the complexity of $\mathcal{O}(T/C \times C^2 \times d)=\mathcal{O}(TCd)$, which is linear in $T$ with constant $C$. If $C > d$, we can re-arrange the order of matrix multiplications \cite{peng2020random} to further reduce its computational complexity:
\begin{equation}
F_c^{\text{spe}}=\underbrace{(Q_c K_c^{T})}_{\mathbb{R}^{C\times C}}V_c^{\text{spe}} \xrightarrow{\text{~}}F_c^{\text{spe}}=Q_c \underbrace{\left(K_c^{T}V_c^{\text{spe}}\right)}_{\mathbb{R}^{d\times d}} ,
\end{equation}
where the re-arranging computation reduces the self-attention complexity in each chunk from $\mathcal{O}(C^2d)$ to $\mathcal{O}(d^3)$.

% \begin{figure}[t]
% \centering
% \includegraphics[width=0.9\columnwidth]{BLACE.pdf}
% \caption{\textbf{(Top)} Proposed linear self-attention with global and local components with top-2 selection strategy, \textbf{(Bottom)} Quadratic attention. Our method significantly reduces the complexity of quadratic attention (gray links).}
% \label{blace_f}
% \end{figure}

\textbf{Modality-shared Attention}. In the ACSR task, lip and hand are complementary with each other to convey the same semantic knowledge, which is more effective to handle the similar labial shapes of lip reading (\textit{e.g.}, \textit{[p]} and \textit{[b]}). Modality-shared attention aims to fuse the important information among different modalities to further alleviate such visual ambiguity. The core idea is to remove the redundant tokens within each chunk and compute modality-shared attention over the remaining vital tokens. Motivated by Theorem \ref{theorem1}, SVD decomposition can be utilized for low-rank approximation of the attention matrix to focus on the most important part of each modality, but will introduce additional complexity. Alternatively, we propose a novel token utilization rate (TUR) to avoid the additional complexity of the SVD decomposition.

\begin{definition}
\label{turd}
    (\textbf{Token Utilization Rate}) Let $A_i^{\text{spe}} \in \mathbb{R}^{C\times C}$ be the modality-specific attention matrix for $i$-th local chunk, and let $C_{i}^j$ denote the $j$-th token of $i$-th chunk. Then, the utilization rate for $C_{i}^j$ is defined as
    \begin{equation}
        \mathbf{TUR(i,j)}= \frac{\sum_{m\neq j}^CA_i^{\text{spe}}(m, j)}{A_i^{\text{spe}}(j, j)}.
    \end{equation}
\end{definition}

As an essential concept, TUR reflects the importance degree of a token for representing all other tokens. When $\text{TUR}(i,j)$ is close to 0, $j$-th token almost only attends itself in the self-attention computation, which implies that other tokens can be represented by the linear combination of the whole sequence except for $j$-th token. This means that the $j$-th token is less critical to formulating the attention score matrix. Conversely, larger $\text{TUR}(i,j)$ indicates that $j$-th token is necessary to represent all other tokens during self-attention formulation, \textit{i.e.}, the span space involving $j$-th token is informative to represent other tokens via linear combinations.

According to Definition \ref{turd}, we select top-$k$ tokens with the highest $\text{TUR}$ values within each chunk in $K$ and $V$ respectively, called TUR-based top-$k$ selection. which reduces the length dimension from $T$ to $CK$. $k$ is the hyper-parameter. Then we compute a $(T \times Ck)$-dimensional attention matrix via the scaled dot-product operation:
\begin{equation}
F^{\text{sha}} = A^{\text{sha}}\hat{V}^{\text{sha}}=\underbrace{\psi(Q \hat{K}^{T})}_{\mathbb{R}^{T\times Ck}} \hat{V}^{\text{sha}},
\end{equation}
where $\hat{K}, \hat{V} \in \mathbb{R}^{Ck\times d}$ denotes the selected key, value vectors. This formulation only requires $O(kTC)$ time and space complexity. Thus, if choosing a very small sampling frequency $k$, such that $k \ll T$, we can significantly reduce the memory and space consumption. 

Here, we additionally define a chunk utilization rate (CUR), which can reflect the importance degree of a chunk in the whole sequence. In the experiment section, we will show the distributions of TUR and CUR to indicate the effectiveness of the proposed method.
\begin{definition}
\label{dcur}
   (\textbf{Chunk Utilization Rate}) Given a sequence with $P$ chunks, $C$ denotes the token number within a chunk. Then, the chunk utilization rate for $i$-th chunk is defined as
    \begin{equation}
        \mathbf{CUR(i)}= \frac{\sum_{j}^C \mathbf{TUR(i ,j)}}{\sum_{i}^P\sum_{j}^C \mathbf{TUR(i ,j)}}.
    \end{equation}
\end{definition}

\subsection{Multi-modal Fusion for Modality-shared Attention}
\label{multimodal}
\begin{figure}[!t]
\centering
\includegraphics[width=1\columnwidth]{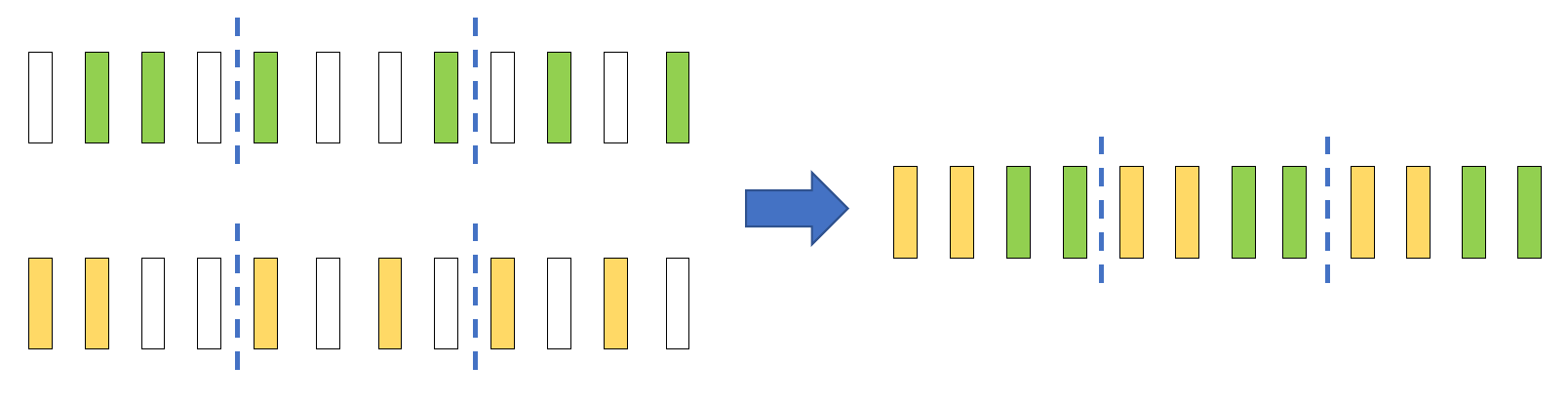}
\caption{An example of the multi-modal fusion. We concatenate the important tokens of different modalities and then calculate the modality-shared attention.}
\label{fusion_fig}
\end{figure}

For multi-modal transformers, the dominant paradigm of cross-modal interaction mainly relies on the cross-attention mechanism \cite{chen2021crossvit}, which still suffers from quadratic complexity and requires additional computations for Q/K/V projections. Therefore, it remains a significant challenge to effectively integrate the important modality information while preserving an efficient manner for the ACSR. In this part, we present a flexible multi-modal fusion strategy based on the TUR. The idea is to force attention flow over important tokens of different modalities within a layer as shown in Figure \ref{fusion_fig}.

Given the token sequences $Q^0, K^0, V^{\text{sha-0}}$ for modality $m_0$ and $Q^1, K^1, V^{\text{sha-1}}$ for modality $m_1$, the first step is to compute the cross-modal K/V vectors via the chunk-level fusion:
\begin{equation}
\begin{aligned}
K^\text{fsn} &=\operatorname{concat}([K^0_0, K^1_0], \cdots, [K^0_{n-1}, K^1_{n-1}]),\\
V^\text{fsn} &=\operatorname{concat}([V_0^{\text{sha-0}}, V_0^{\text{sha-1}}], \cdots, [V_{n-1}^{\text{sha-0}}, V_{n-1}^{\text{sha-1}}]). 
\end{aligned}
\label{fusion}
\end{equation}
Then the modality-shared attention for cross-modal interaction is formulated as follows:
\begin{equation}
\begin{aligned}
    &F_0^{\text{sha}} = \psi[Q^0 (K^\text{fsn})^{T}] V^{\text{fsn}},\\ &F_1^{\text{sha}} = \psi[Q^1 (K^\text{fsn})^{T}] V^{\text{fsn}}.
\end{aligned}
\end{equation}
Concretely, we exploit modality-specific query and modality-shared fused key/value vectors, enhancing cross-modal interaction by allowing free attention flows over sequences of different modalities. Note that the above-mentioned concatenation operation induces the double length with respect to the input sequence. To tame the higher quadratic complexity of pairwise attention over double length, the TUR-based top-$k$ selection can be adopted to replace $K^m_i, V^{\text{sha-m}}_i$ with $\hat{K}^m_i, \hat{V}^{\text{sha-0}}_i$ in Eq. \ref{fusion}, where $m \in \{0, 1\}$, which allows to only exchange important information for different modalities via the tokens with higher TUR values.

\subsection{Convolution-based Aggregation}
In this section, we present a ConAgg module to enhance the spatial relations for modality-specific and modality-shared components. Convolution is the default method since it can potentially improve the representative capacity with a limited model size. For brevity, the residual connection and dropout are omitted in the formulation.

% \begin{figure}[!t]
% \centering
% \includegraphics[width=0.3\linewidth, angle=90]{convagg.png}
% \caption{The architecture of the ConAgg module. The ConAgg module contains a depth-wise convolution block and a point-wise convolution block, which can enhance the feature interaction along with the spatial dimension.}
% \label{convagg}
% \end{figure}

\textbf{Addition Merge.} Given the output of the modality-specific component $F^{\text{spe}}$ and the output of the modality-specific component $F^{\text{sha}}$, we add them along the temporal dimension to obtain the final attention output $F_{o} = F^{\text{spe}} + F^{\text{sha}}$.

% and then project them into the output space following the projection analogous to Eq. \ref{gmp}:
% \begin{equation}
%     F_{o} = \phi\{[(F^{\text{spe}} + F^{\text{sha}})\odot G_{u}]W^{o}\}.
% \end{equation}

 \begin{figure}[!t]
\centering
\subfigure[Gated Linear Unit]{\includegraphics[width=0.49\columnwidth]{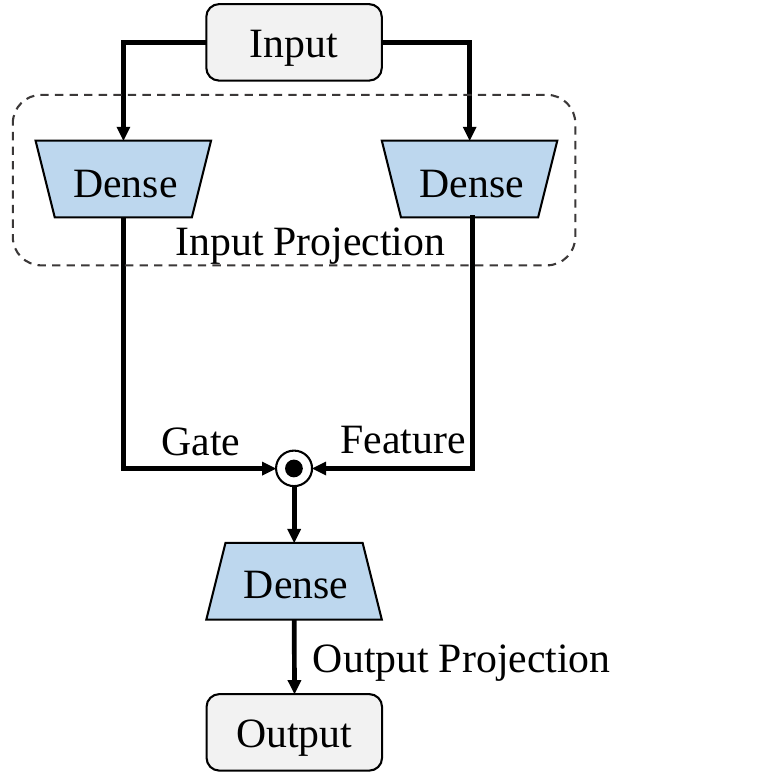}}
\subfigure[Gated Hidden Projection]{\includegraphics[width=0.49\columnwidth]{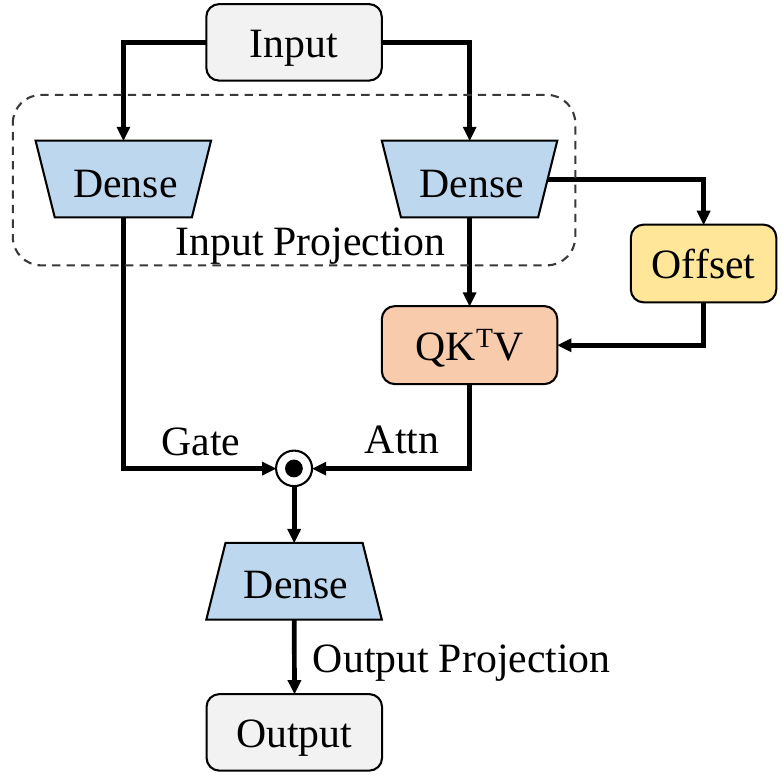}}
\caption{(a) Gated Linear Unit \cite{dauphin2017language}. (b) The proposed Gated Hidden Projection is to control the information flow for TIAA, where each projection is followed by a layernorm layer and an activation function.}
\label{gate_unit}
\end{figure}

\textbf{Spatial Aggregation.} The self-attention explicitly focuses on exploring temporal dependency but less emphasizes spatial relationships. To mitigate this issue, a ConAgg module is utilized to enhance the spatial interaction along with the spatial dimension, which is flexible and insensitive to the input length. More specifically, given the input sequence $F_{o} \in \mathbb{R}^{T\times d_m}$, a depth-wise convolution (DWC) block is exploited to capture local correlations over spatial dimension:
\begin{equation}
\begin{aligned}
    Z_t &= \operatorname{Transpose}(F_{o}) \in \mathbb{R}^{d_m\times T}, \\
    Z_d &= \operatorname{Swish}(\operatorname{BatchNorm}(\operatorname{DWC}(Z_t))),
\end{aligned}
\end{equation}
which adopts Swish \cite{ramachandran2017searching} as the activation function. Then a point-wise convolution (PWC) for the feature projection is to calculate the output $Z_o$ as follows:
\begin{equation}
\begin{aligned}
    Z_p &= \operatorname{Swish}(\operatorname{BatchNorm}(\operatorname{PWC}(Z_d))),  \\
    Z_o &= \operatorname{Transpose}(Z_{p}) \in \mathbb{R}^{T\times d_m}.
\end{aligned}
\end{equation}
The computational costs of the depth-wise convolution are $\mathcal{O}(TDd_m)$, where $D$ is the kernel size. The point-wise convolution has $\mathcal{O}(Td_m)$ complexity. The overall complexity is linear with respect to $T$ with a constant factor $D$.

 % replacing the feed-forward network to further improve the expressiveness of the EcoCued model. 
\subsection{Gated Hidden Projection} 
\label{GHP}
Gated hidden projection is to control the feature flow through the TIAA module as the regularization \cite{dauphin2017language}, replacing the feed-forward network to improve the capacity and flexibility of the EcoCued model. Note that gated hidden projection only contains two fully-connected layers, which is more light-weight than the feed-forward network. The main architecture of gated hidden projection is illustrated in Figure \ref{gate_unit}. 

Given the input sequence $F \in \mathbb{R}^{T\times d_m}$ of the length $T$, transformer’s input projection is formulated as $F_{u} = \phi(FW^{u})$ to obtain the hidden embedding $F_{u}\in \mathbb{R}^{T\times d}$. The output projection is formulated as $\hat{F} =\phi(F_{o} W^{o})$, where $W^{u} \in \mathbb{R}^{d_m\times d}$, $W^{o} \in \mathbb{R}^{d\times d_m}$ and $\phi$ is an element-wise activation function. Here, $F_{o}$ is the TIAA output. Inspired by the augmented MLP \cite{dauphin2017language}, the gate mechanism can control the information flows from the input to output projection, which utilizes a Gated Linear Unit \cite{dauphin2017language} for the input and output projection as:
\begin{equation}
    [F_{u}|G_{u}] = \phi(FW^{u}), \quad \hat{F} = \phi((F_{o}\odot G_{u})W^{o}),
    \label{gmp}
\end{equation}
where the input projection is augmented by $W^{u} \in \mathbb{R}^{d_m\times 2d}$, \textit{i.e.}, providing hidden feature $F_{u}\in \mathbb{R}^{T \times d}$ and gating weight $G_{u} \in \mathbb{R}^{T \times d}$. Here $[\cdot|\cdot]$ denotes the chunk operation, and $\odot$ stands for element-wise multiplication. In this case, the output representations $\hat{F}$ are gated by the weight $G_{u}$, which are associated with the same input projection, enabling higher computing efficiency combined with the self-attention mechanism. The final output can be added with the original input as a residual connection.

\begin{table*}[!t]
\caption{The details of CS datasets with different languages. The \#Train/\#Test is in the form of Sentences/Characters. The \#Cuer is in the format of people number and type, \textit{i.e.}, hearing (H) or hearing-impaired (HI) people}.
\begin{center}
\begin{tabular}{c|c|c|c|c|c|c}
\hline
Dataset    & French & \multicolumn{2}{c|}{British}	& \multicolumn{3}{c}{Chinese}	\\ \hline
\#Cuer & 1-HI & 1-HI	& 5-HI	& 1-H & 4-H & 1-HI    \\ \hline
\#Sentence & 238 & 97 & 390	  &1000   & 4000 & 818\\ \hline
\#Character & 12872 & 2741 & 11021	  &32902   & 131581 & 25244\\ \hline
\#Word & - & - & -	  & 10562   &  42248 & 8269\\ \hline
\#Phoneme & 35 & 44 & 44 & 40 & 40 & 40\\ \hline
\#Shape & 8 & 8 & 8 & 8 & 8 & 8\\ \hline
\#Position & 5 & 4 & 4 & 5 & 5 & 5\\ \hline
\#Train & 193/10636 & 78/2240 & 312/8924 & 800/26683 & 3200/105372 &652/20209\\ \hline
\#Test & 45/2236 & 19/501 & 78/2097 & 200/6219 & 800/26209 & 166/5035\\ \hline
\end{tabular}
\begin{tablenotes}
\centering
\item[1] \small * X-H (HI) denotes X hearing (hearing-impaired) cuers, where X is the number of cuers.
\end{tablenotes}
\end{center}
\label{dataset}
\end{table*}

 \begin{table*}[!t]
\caption{Performance comparison with baselines on Chinese CS dataset. The chunk size is 32 and $k$ is 4. Bold denotes the best results. The inference time is measured using a $(1, 100, 3, 64, 64)$ tensor. The VPS denotes the processed video number per second.}
\begin{center}
\begin{tabular}{l|c|c|c|c|c|c|c|c}
\hline
\multicolumn{2}{c|}{Method} & \multicolumn{4}{c|}{Chinese}    & \multicolumn{3}{c}{\multirow{2}{*}{Speed Up}} \\ \cline{1-6}
\multicolumn{2}{c|}{\#Cuer}  & \multicolumn{2}{c|}{single}    & \multicolumn{2}{c|}{multiple}  &\multicolumn{2}{c}{} \\ \hline
Metrics           & Param(M)& CER   & WER   & CER   & WER     & Inference Time (ms) & VPS  & FLOPs   \\ \hline
ResNet18 \cite{he2016deep}         & 11.7    & 35.6  & 78.3  & 41.9  & 83.4    &  46.73   & 21.39  & 56.63G  \\ \cline{2-9}
 + LSTM  \cite{papadimitriou2021fully}         & 22.7    & 55.4  & 92.8  &  61.4 & 96.1    &  49.35   & 20.26 & 149.50G   \\ \hline
JLF      \cite{wang2021cross}         & $<$1    & 33.5  & 67.1  & 68.2  & 98.1    &  12.64  &  79.11 & 8.16G        \\ \hline
CMML     \cite{liu2023cross}         & 54.9    & 9.7   & 24.1  & 24.5  & 54.5    &  52.47  & 19.06 & 156.06G  \\ \hline
CNN + MHSA  \cite{vaswani2017attention}      & 29.3    & 26.1  & 61.8  & 38.8  & 78.6    &  50.63  & 19.75  & 109.01G \\ \hline
CNN + FLASH \cite{hua2022transformer}      & 7.7     & 36.4  & 75.2  & 43.4  & 83.9    &  48.59  & 20.58 & 13.66G  \\ \hline
CNN + Linformer \cite{wang2020linformer}  & 8.9     & 39.3  & 79.7  & 42.9  & 81.1    &  47.26  & 21.16  & 17.72G  \\ \hline
CNN + Performer \cite{choromanski2020rethinking}  & 11.0    & 32.1  & 71.4  & 44.4  & 82.6    &  47.87  & 20.88 & 15.31G \\ \hline
CNN + Cosformer \cite{qin2021cosformer}  & 7.2     & 30.6  & 74.7  & 41.2  & 79.5    &  46.64  & 21.44  & 13.58G \\ \hline
Ours (Random)     & 6.6     & 23.2  & 56.2  & 31.8  & 68.9    &  45.14  &  22.15 & 13.15G\\ \hline
Ours (TUR)             & \textbf{6.6}  & \textbf{9.0}   & \textbf{24.1}  &  \textbf{22.2}   &  \textbf{53.8} &  \textbf{45.14} &  \textbf{22.15} &    \textbf{13.15G}\\ \hline
\end{tabular}
\end{center}
\label{acsr}
\end{table*}

\section{Experiments}
\subsection{Experimental Setup}
\noindent\textbf{Datasets.} We use three public benchmarks to evaluate the performance of the proposed method, \textit{i.e.}, the Mandarin Chinese \cite{liu2023cross,gao2023novel,liu2022objective}, French \cite{liu2018automatic}, and British English \cite{sankar2022multistream}. The Mandarin Chinese CS dataset is the first large-scale multi-cuer CS benchmark for Mandarin Chinese, including 4,000 sentences for 4 cuers. Chinese vowels and consonants are categorized by 40 phonemes, represented by hands (8 shapes and 5 positions) and corresponding lips. Both British English and French CS have a single-cuer setting with 97 and 238 sentences, respectively. Multi-cuer data of the British English CS dataset is not open-sourced. In detail, 35 French phonemes are represented by hands (8 shapes and 5 positions) and corresponding lips, while 8 hand shapes and 4 hand positions for British English. The training and test sentences are randomly split as $4:1$ without repeated sentences. For the data pre-processing, two open-source packages are used to segment the ROIs from the lip and hand videos, \textit{i.e.}, dlib and mediapipe\footnote{\textbf{dlib:} http://dlib.net, \textbf{mediapipe:} https://mediapipe.dev}. For all datasets, the frame per second (FPS) of videos is 30. Each video is annotated by a sentence text instead of frame-wise labels used by most previous ACSR methods. Phoneme-level classification is required for the training and inference for the sequence-to-sequence task. Besides, we collect 818 Chinese CS sentences with videos recorded by one hearing-impaired cuer to further verify the effectiveness of the proposed method. Such a setting is challenging for the ACSR task due to ambiguous lip-reading and faster hand movements with blurring. More details of public CS datasets can refer to Table \ref{dataset}. 

\noindent\textbf{Implementation Details.} We utilize Pytorch to implement the whole learning framework. One Nvidia V-100 GPU is used for all experiments. For the input videos, each frame is resized to $64 \times 64$. RandAugment \cite{cubuk2020randaugment} is utilized as the augmentation of the training data. During training, the EcoCued is randomly initialized. RegNet \cite{Radosavovic_2020_CVPR} is used as the front-end backbone for all baselines, which is initialized using pre-trained weights on ImageNet. The EcoCued contains $3$ TIAA layers, where the other settings are the same as \cite{vaswani2017attention}. The Adam optimizer with $\beta_1 = 0.9$, $\beta_2 = 0.999$ and $\epsilon= 0.05$ is used for end-to-end training. The mini-batch size is set as 1. $d_m$ is 256 and $d$ is 64. The learning rate increases linearly with the first $5,000$ steps, yielding a peak learning rate, and then decreases proportionally to the inverse square root of the step number. The whole network is trained for $50$ epochs. 

\noindent\textbf{Evaluation Metric.} (1) To demonstrate the effectiveness of the proposed method, we utilized several previous ACSR solutions as the comparisons including ResNet18 + CTC \cite{he2016deep}, ResNet18 + LSTM \cite{papadimitriou2021fully}, JLF \cite{wang2021cross}, and CMML \cite{liu2023cross}. CMML is the previous SOTA method. The transformer methods are also involved. In detail, the vanilla Multi-Head Self-Attention (MHSA) \cite{vaswani2017attention} is included as a standard baseline. Further transformers with lower complexity are involved as stronger baselines involving FLASH \cite{hua2022transformer}, Linformer \cite{wang2020linformer}, Performer \cite{choromanski2020rethinking}, and Cosformer \cite{qin2021cosformer}. (2) To evaluate the effectiveness of the TUR strategy, we utilized a random token selection strategy as a baseline. (3) To evaluate the generalization to other multi-modal task, we also conducted the comparison experiments on the audio-visual speech recognition task. (4) To verify the generalization of our method, we conducted the comparison experiments on LRS2-BBC dataset \cite{afouras2018deep} for audio-visual speech recognition. All approaches are evaluated using character error rate (CER) and word error rate (WER) to indicate the ACSR recognition ability on both phoneme and word levels.

\begin{table}[!t]
\caption{Performance comparisons on hearing-impaired people on Chinese CS dataset. CMML is the previous SOTA.}
\begin{center}
\begin{tabular}{c|c|c}
\hline
Method & CER & WER  \\ \hline
ResNet18 + MHSA \cite{papadimitriou2021fully} & 65.1 & 97.8  \\ \hline
CMML \cite{liu2023cross} &  32.0   &  67.0   \\ \hline
Ours                     &  \textbf{29.5}  &  \textbf{61.5} \\ \hline
\end{tabular}
\end{center}
\label{disable}
\end{table}
\vspace{-0.5cm}

\begin{figure*}[!t]
\begin{center}
\includegraphics[width=0.7\linewidth]{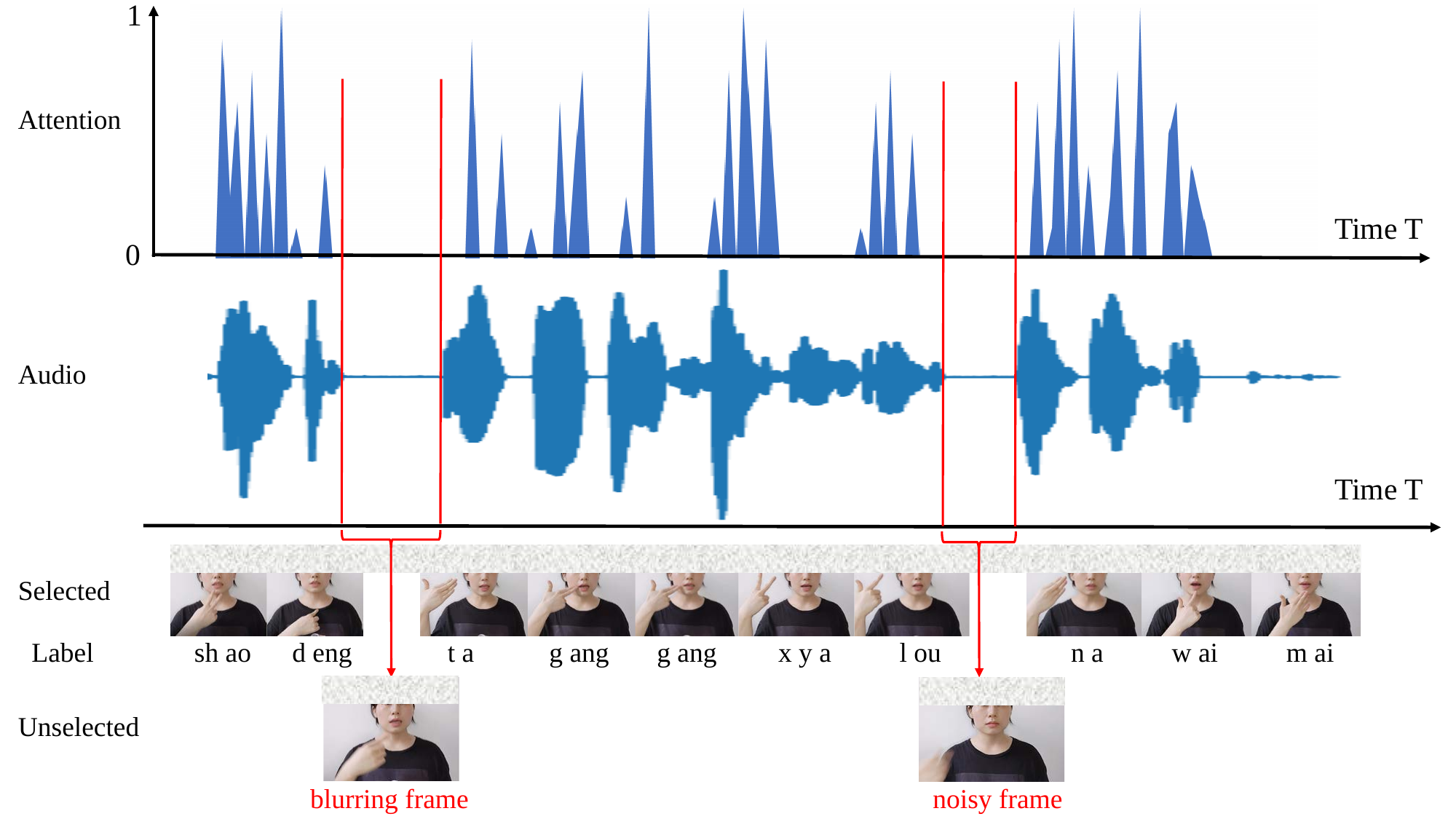}
\caption{Visualization of modality-shared attention for the selected tokens on the Chinese CS dataset. X-axis denotes the time. Y-axis denotes the attention score (first row), audio (second row), and corresponding frames in the videos (third and fourth rows). The attentions of the selected tokens are well aligned with the audio signal distribution of the cuer. Besides, we can see that the selected tokens mainly correspond to the video frames with less visual ambiguity.}
\label{vis_index}
\end{center}
\end{figure*}

\subsection{Compared with Previous Methods}
\noindent\textbf{Chinese CS Dataset.} In Table \ref{acsr}, we present the results on the Chinese CS dataset for hearing people. Both recognition accuracy and parameters are provided to show the effectiveness of the proposed method. As suggested in Table \ref{acsr}, the proposed method achieves significant performance improvement on both CER and WER on all evaluation sets, \textit{i.e.}, 9\% CER and 24.1\% WER on the single cuer setting, as well as 22.2\% CER and 53.8\% WER on the multiple cuer setting. Also, the previous SOTA method CMML obtained good performance using about $54.9M$ parameters, while our method only utilizes $6.6M$ parameters to achieve similar results. Table \ref{acsr} also presents the comparisons with recent linear transformers. Our EcoCued performs superior results compared with them, even outperforms the vanilla Transformer \cite{vaswani2017attention}. Besides, we notice that previous linear transformers have a significant performance drop on the ACSR task. The main reason lies in that they may drop some important information due to the accumulated approximation errors \cite{wu2022flowformer} and lack of effective cross-modal fusion strategies, while our method requires modeling attentive information on the important tokens and achieves a flexible fusion for multi-modal inputs. Additionally, our method exhibits faster inference speed than other efficient transformers.

As shown in Table \ref{disable}, our method can also achieve the best results on the Chinese CS data of hearing-impaired people. Compared with CMML and vanilla self-attention, our method can further improve performance via the more effective and flexible cross-modal interaction. Compared with hearing people, CS data of hearing-impaired people is more challenging for the applications of the ACSR model. For example, there may exist visual ambiguity in the hand shapes because hands may move fast with blurring. 
%Hence the asynchronous effect between lips and hands is more significant. 
Besides, the lip reading performance of hearing-impaired people is slightly more ambiguous than the hearing ones. Thus, the performance of hearing-impaired people is still relatively lower than that of hearing people.

\begin{table}[!t]
\caption{Performance comparisons (CER) on British and French CS datasets. WER is unavailable due to lacking word-level annotations. CMML is the previous SOTA.}
\begin{center}
\begin{tabular}{c|c|c}
\hline
Dataset & French & British  \\ \hline
\#Cuer & single                & single  \\ \hline
ResNet18 + HMM \cite{liu2018visual}    & 38.0 &-  \\ \hline
ResNet18 + LSTM \cite{papadimitriou2021fully} & 33.4 & 43.6 \\ \hline
ResNet18 + MHSA \cite{papadimitriou2021fully} & 37.5 & 39.8  \\ \hline
Student CE \cite{wang2021cross}    &  35.6   & 47.5   \\ \hline
JLF1 \cite{wang2021cross}         &  27.5   & 38.5   \\ \hline
JLF2 \cite{wang2021cross}          & 27.5   & 36.9     \\ \hline 
JLF3 \cite{wang2021cross}         &  25.8   & 35.1    \\ \hline
CMML \cite{liu2023cross}              &  24.9   &  33.6   \\ \hline
Ours  &  \textbf{24.8}  &  \textbf{33.0} \\ \hline
\end{tabular}
\end{center}
\label{BCS}
\end{table}

\noindent\textbf{French\&British CS Dataset.} As shown in Table \ref{BCS}, our method can achieve the best results on both French and British CS datasets. Compared with LSTM and vanilla transformer, our method benefits from the effective cross-modal interaction and can capture long-time dependency over multi-modal data streams. The accuracy improvement is slight due to the small data scale of these datasets. Besides, our method can outperform the previous SOTA method CMML, which indicates that the proposed efficient method can achieve competitive performance with lower model complexity. 

\begin{table}[!t]
    \vspace{-0.5cm}
    \caption{Performance comparison with baselines on LRS2-BBC dataset.}
    \begin{center}
    \begin{tabular}{l|c}
    \hline
    Method & WER \\ \hline
    TM-CTC \cite{afouras2018deep}          & 16.70   \\ \hline
    TM-Seq2Seq \cite{afouras2018deep}      & 8.5  \\ \hline
    TDNN \cite{yu2020audio}            & 5.90    \\ \hline
    CNN + Conformer \cite{ma2021end}   & 4.20    \\\hline
    Ours                & \textbf{3.95}    \\\hline
    \end{tabular}
    \end{center}
    \label{lrs2_bbc}
    \vspace{-0.5cm}
\end{table}

\noindent\textbf{LRS2-BBC Dataset.} We also conducted the comparison experiments for the audio-visual speech recognition (AVSR) task. As shown in Table \ref{lrs2_bbc}, compared with vanilla transformers with CTC and Seq2Seq decoding, our method can achieve better results on LRS2-BBC dataset, indicating the generalization of the proposed efficient method to the AVSR task.
    
% \begin{table}[!t]
%  \caption{Complexity analysis for EcoCued and recent efficient self-attention techniques. Although there are some works with $\mathcal{O}(T)$ complexity (\textit{i.e.}, \cite{hua2022transformer,wang2020linformer,choromanski2020rethinking,qin2021cosformer}), they perform worse for the ACSR task due to the lack of the flexible and effective multi-modal fusion (their performance can be seen in Table \ref{acsr}).}
% \begin{center}
% \small
% \begin{tabular}{l|c}
% \hline
% Method & Complexity\\ \hline
% CMML \cite{liu2023cross}      & $\mathcal{O}(T^2)$     \\ \hline
% MHSA \cite{vaswani2017attention}            &  $\mathcal{O}(T^2)$      \\ \hline
% FLASH  \cite{hua2022transformer}            &  $\mathcal{O}(T)$        \\ \hline
% Linformer \cite{wang2020linformer}          &  $\mathcal{O}(T)$        \\ \hline
% Performer  \cite{choromanski2020rethinking} &  $\mathcal{O}(T)$        \\ \hline
% Cosformer  \cite{qin2021cosformer}          &  $\mathcal{O}(T)$        \\ \hline
% Ours       &  $\mathcal{O}(T)$        \\ \hline
% \end{tabular}
% \end{center}
% \label{com}
% \end{table}

\noindent\textbf{Computational Complexity Analysis.} 
In this part, we analyze the computational complexity of the proposed method and recent efficient self-attention techniques. Standard self-attention \cite{vaswani2017attention} and CMML \cite{liu2023cross} calculate the full pair-wise attention in the sequence, resulting in $\mathcal{O}(T^2)$ complexity. Linformer \cite{wang2020linformer} adopts two linear projections to shrink the length dimension, leading to complexity $\mathcal{O}(T)$. FLASH \cite{hua2022transformer} introduces cumsum operation to reduce the cost of auto-regressive with complexity $\mathcal{O}(T)$. Performer \cite{choromanski2020rethinking} adopts kernelizable attention to approximate the softmax operation with complexity $\mathcal{O}(T)$. Cosformer \cite{qin2021cosformer} replaces the softmax operation with a linear function with complexity $\mathcal{O}(T)$. Importantly, due to a lack of effective multi-modal fusion with spatial-temporal interactions, these transformers (\textit{i.e.}, \cite{hua2022transformer,wang2020linformer,choromanski2020rethinking,qin2021cosformer}) with linear complexity $\mathcal{O}(T)$ exhibit worse performance on the ACSR task. Our method decomposes the full attention into modality-specific and modality-shared components with complexity $\mathcal{O}(T)$, which can capture both long-time temporal dependencies and spatial relations for different modalities. 

Additionally, we provide the comparisons for speed-ups of inference, including inference time, VPS, and FLOPs. FLOPs is used to compute the number of operations for a given model. VPS indicates that how many videos that the model can process in one second. As shown in Table \ref{acsr}, it is observed that the inference speed of our method is faster than the most baselines over all protocols. Although JLF achieves the fastest inference time, it obtains the worse recognition accuracy.

% \begin{figure}[ht]
% \vskip 0.2in
% \begin{center}
% \centerline{\includegraphics[width=\columnwidth]{chunk_ur.pdf}}
% \caption{Histograms of chunk utilization rates of lips and hands trained by our model.}
% \label{icml-historical}
% \end{center}
% \vskip -0.2in
% \end{figure}

% \begin{algorithm}[tb]
%    \caption{Bubble Sort}
%    \label{alg:example}
% \begin{algorithmic}
%    \STATE {\bfseries Input:} data $x_i$, size $m$
%    \REPEAT
%    \STATE Initialize $noChange = true$.
%    \FOR{$i=1$ {\bfseries to} $m-1$}
%    \IF{$x_i > x_{i+1}$}
%    \STATE Swap $x_i$ and $x_{i+1}$
%    \STATE $noChange = false$
%    \ENDIF
%    \ENDFOR
%    \UNTIL{$noChange$ is $true$}
% \end{algorithmic}
% \end{algorithm}

% \begin{table}[t]
% \caption{Classification accuracies for naive Bayes and flexible
% Bayes on various data sets.}
% \label{sample-table}
% \vskip 0.15in
% \begin{center}
% \begin{small}
% \begin{sc}
% \begin{tabular}{lcccr}
% \toprule
% Data set & Naive & Flexible & Better? \\
% \midrule
% Breast    & 95.9$\pm$ 0.2& 96.7$\pm$ 0.2& $\surd$ \\
% Cleveland & 83.3$\pm$ 0.6& 80.0$\pm$ 0.6& $\times$\\
% Glass2    & 61.9$\pm$ 1.4& 83.8$\pm$ 0.7& $\surd$ \\
% Credit    & 74.8$\pm$ 0.5& 78.3$\pm$ 0.6&         \\
% Horse     & 73.3$\pm$ 0.9& 69.7$\pm$ 1.0& $\times$\\
% Meta      & 67.1$\pm$ 0.6& 76.5$\pm$ 0.5& $\surd$ \\
% Pima      & 75.1$\pm$ 0.6& 73.9$\pm$ 0.5&         \\
% Vehicle   & 44.9$\pm$ 0.6& 61.5$\pm$ 0.4& $\surd$ \\
% \bottomrule
% \end{tabular}
% \end{sc}
% \end{small}
% \end{center}
% \vskip -0.1in
% \end{table}

\subsection{Effectiveness of TUR-based Top-$k$ Selection}
\begin{figure}[!t]
\begin{center}
\includegraphics[width=1\columnwidth]{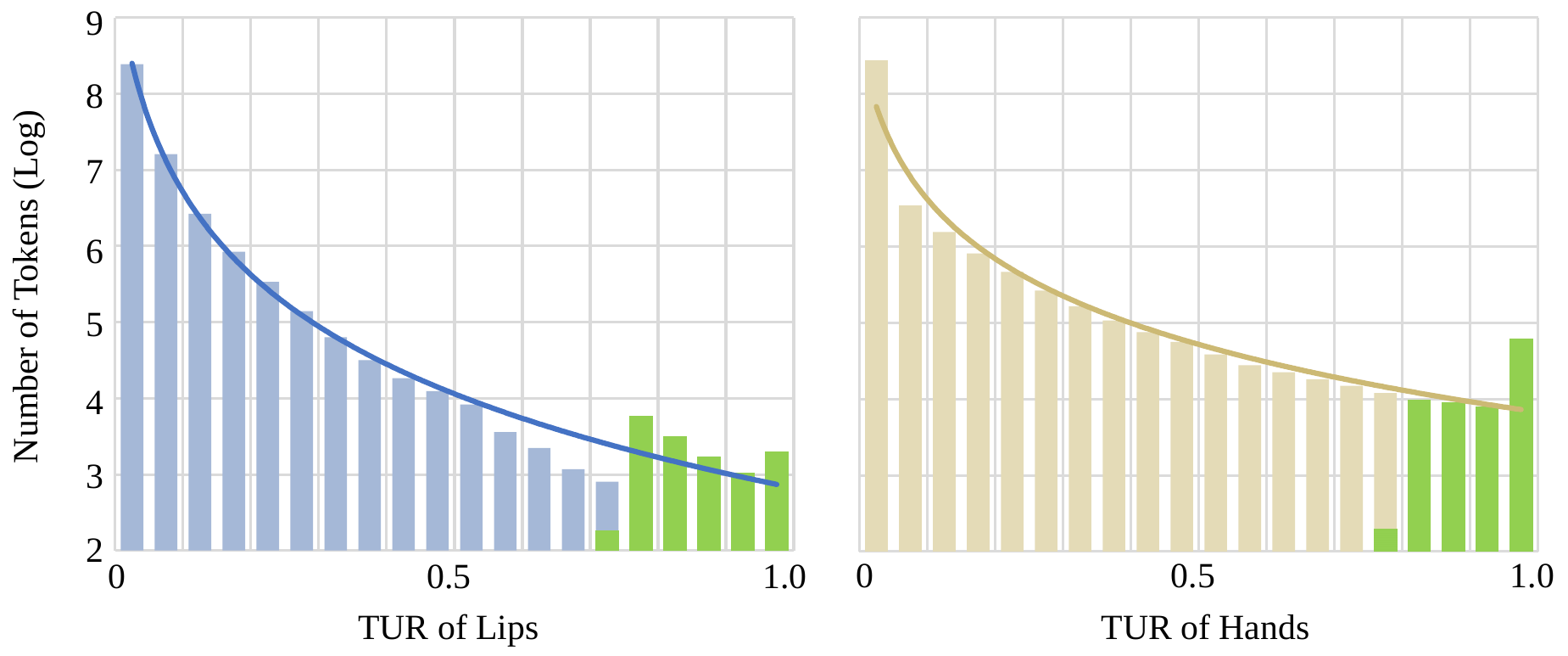}
\caption{TUR Histograms of lips (left) and hands (right) using our model. The green part is for the selected tokens with higher TUR values. TURs are normalized within each chunk. Different modalities exhibit similar TUR distributions and most tokens perform small TUR values. We can see that most tokens perform small TUR values, and our EcoCued can maintain the important tokens with higher TUR values. }
\label{tur}
\end{center}
\end{figure}

\begin{figure}[!t]
\begin{center}
\subfigure[Without Top-$k$ Selection]{\includegraphics[width=1\linewidth]{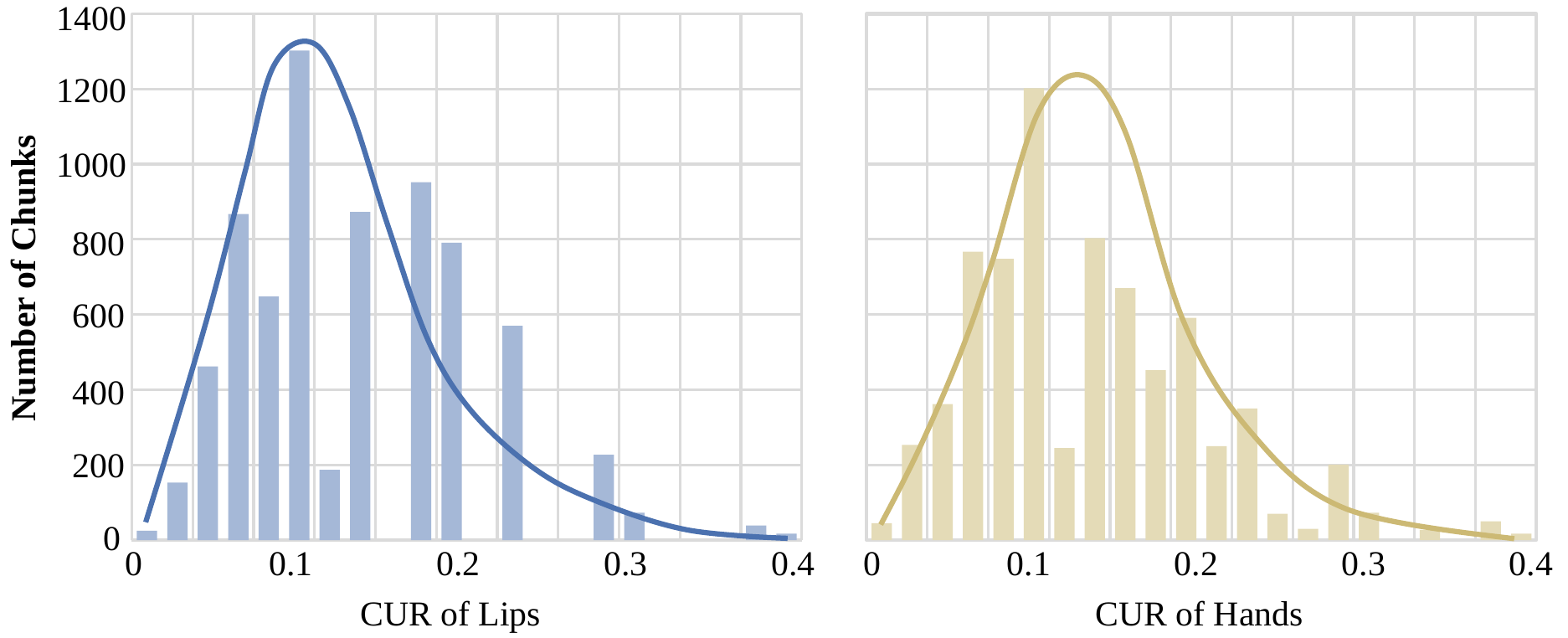}}
\\
\subfigure[With Top-$k$ Selection]{\includegraphics[width=1\linewidth]{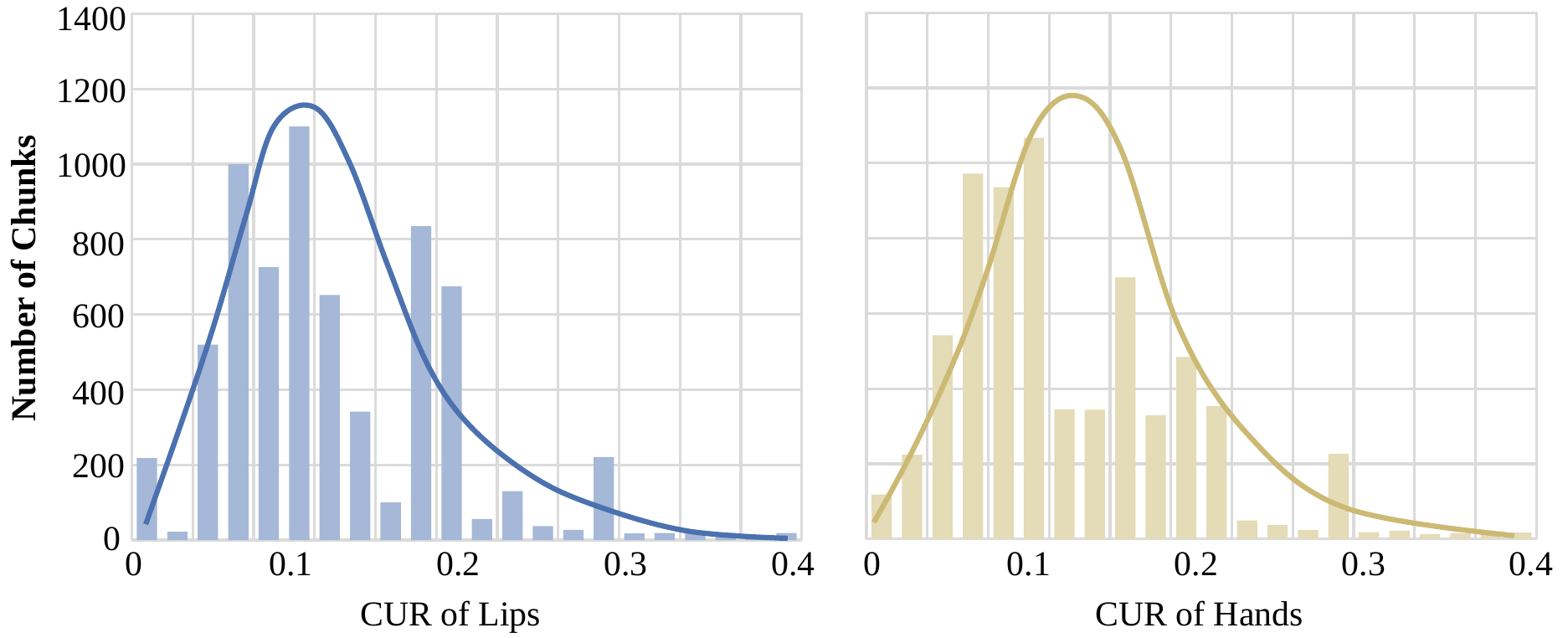}}
\caption{CUR Histograms of lips (left) and hands (right) using our model. Top is without top-$k$ selection and the Bottom is with top-$k$ selection. CURs are normalized within each sequence. Different modalities exhibit skewness CUR distributions. Most tokens perform smaller CUR values.}
\label{v_cur}
\end{center}
\end{figure}

In this part, we study the impact of the TUR-based top-$k$ selection strategy. To this end, we visualize the attention of selected tokens and the distributions of TUR and CUR, as shown in Figure \ref{vis_index}, \ref{tur}, and \ref{v_cur}, respectively.

% \begin{figure}[!t]
% \begin{center}
% \subfigure[Without Top-$k$ Selection]{\includegraphics[width=1\linewidth]{out_tur.pdf}}
% \\
% \subfigure[With Top-$k$ Selection]{\includegraphics[width=1\linewidth]{with_tur.pdf}}
% \caption{Histograms of TUR of lips and hands using our model. Left is without top-$k$ selection and the Right is with top-$k$ selection. TURs are normalized within each chunk. Different modalities exhibit similar TUR distributions and most tokens perform small TUR values.}
% \label{tur}
% \end{center}
% \end{figure}

\textbf{Distribution of Selected Tokens.}
In this part, we exhibit the attention scores for the selected tokens on the Chinese CS dataset in Figure \ref{vis_index}, where attention scores are normalized within each chunk. The red lines indicate the segment where the tokens are not selected with lower TUR values. It is observed that the attentions of the selected tokens (blue spikes in the first row) are well aligned with the audio signal distribution from the cuer (blue spikes in the second row). As shown in the third row, our method mainly pays attention to the video frames with less visual ambiguity caused by hand movements. These tokens are representative with clear hand shapes in the sequence, which further validates the effectiveness of TUR.

% and $k=8, C=32$
\textbf{Distribution of Token Utilization Rate.} We present the TUR distributions on the Chinese CS dataset in Figure \ref{tur}, where TUR values are normalized within each chunk and $k=4, C=32$. As shown Figure \ref{tur}, for both lip and hand modalities, normalized TUR exhibits an exponential distribution, where most tokens have relatively small TUR values. Besides, TUR performs a similar tendency for different modalities. This confirms that many tokens contribute less to the self-attention, resulting in the low-rank property. With top-$k$ selection, our method can preserve the important tokens with the higher TUR values, \textit{i.e.}, green color. Therefore, it is concluded that the top-$k$ selection strategy can maintain the tokens with higher TUR values, significantly reducing the complexity while keeping the performance quality.

\textbf{Distribution of Chunk Utilization Rate.} To validate the TUR's effectiveness, we empirically observe modality-shared attention changes with and without top-$k$ selection. To achieve this, we first define Chunk Utilization Rate (CUR) in Definition \ref{dcur}, then we visualize the CUR distributions on the Chinese CS dataset in Figure \ref{v_cur}, where CUR values are normalized within each sequences and $k=4, C=32$. As shown in Figure \ref{v_cur} (a), normalized CUR exhibits a skewness distribution without top-$k$ selection, which indicates the diverse attention scores of different chunks, $\textit{i.e.}$, different attention matrix ranks for the different chunks. Besides, most chunks have relatively small CUR values, which further confirms the low-rank property of the self-attention. Comparing Figure \ref{v_cur} (a) with (b), CUR curves perform a similar tendency with and without top-$k$ selection, which indicates that the top-$k$ selection maintains the critical global information in a sequence. Importantly, Z test in the statistics area is used to measure the significant difference for CUR values with and without topk selection. Z-value, p-value, and confidence are: $Z=1.645, p=0.0001<0.001, \alpha=0.05$, indicating that there isn't a significant difference for CUR values with and without top-$k$ selection. This keeps the performance quality when TIAA decomposes the full attention.

% due to that CUR values follow the approximately normal distribution, Z test \cite{moore1989introduction} in the statistics area is used to measure the significant difference for CUR values with and without top-$k$ selection. The $p$-value for this test is $99.9\%>5\%$, indicating that there isn't a significant difference for CUR values with and without top-$k$ selection. 

\begin{figure}[!t]
\centering
\includegraphics[width=1\linewidth]{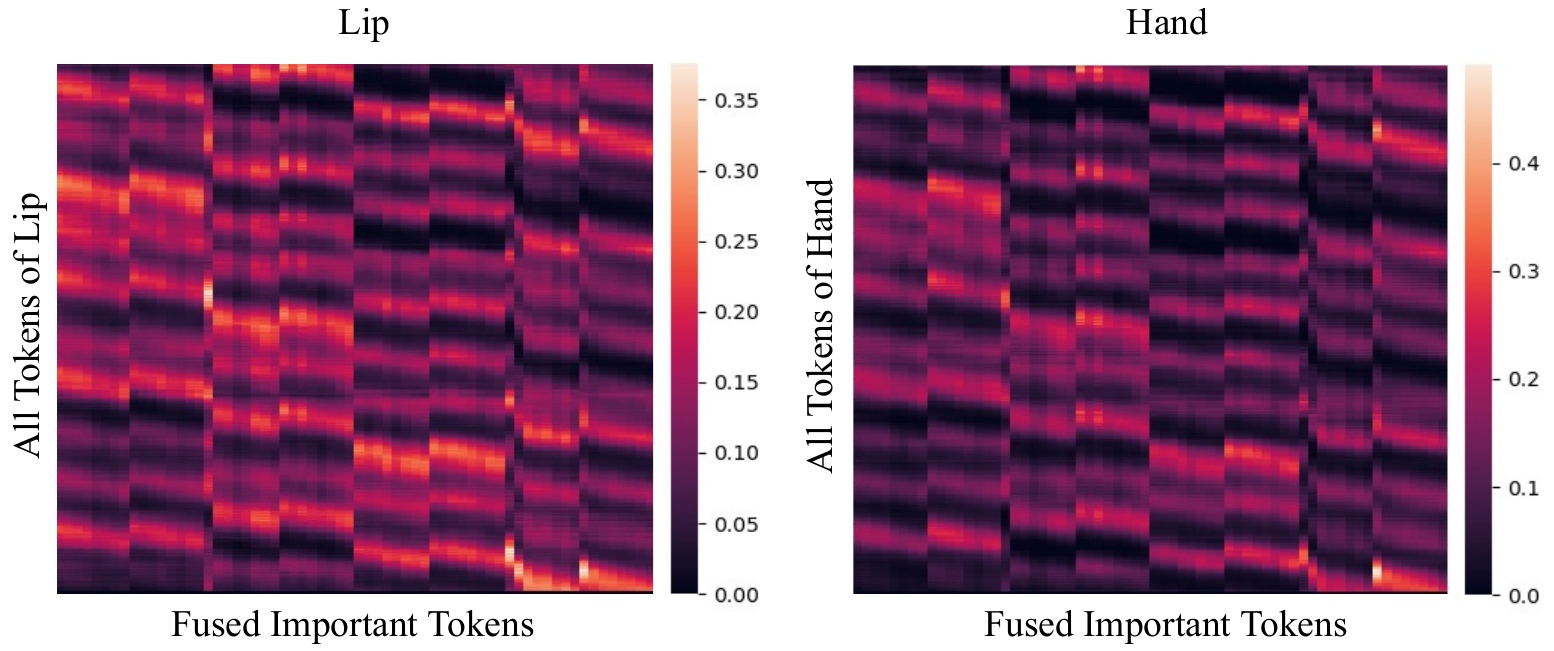}
\caption{Heat maps of modality-shared attentions for lip (left) and hand (right) on Chinese CS dataset. The attention matrices are from the final TIAA layer in the EcoCued for one CS video in the test set.} %图片标题
\label{ms_attn}  %图片交叉引用时的标签
\end{figure}

\textbf{Analysis of the Asynchronous Multi-modal Issue in ACSR.} 
\label{chrony}
In the literature, researchers have investigated the asynchronous phenomenon between lip and hand movements in CS and observed that, during the cuing process, the hand typically reaches its target before lip movements \cite{attina2004pilot, attina2005temporal,liu2020re}. The duration of hand preceding time varies and is cuer-dependent, which makes multi-modal fusion in ACSR more difficult. Indeed, our fusion method in this work does not explicitly address the asynchronous multi-modal issue in ACSR. Instead, we propose an effective computation and parameter-efficient transformer-based fusion method to consider the global dependency over the long sequence inputs of the CS multiple modalities, realizing efficient multi-modal learning.

Given the effectiveness of our method, we believe that it could indirectly alleviate the above-mentioned asynchronous issue.
To demonstrate this point, we show the modality-shared attention score matrix of the TIAA module for lip and hand modalities (see Figure \ref{ms_attn}). We hypothesize that the modality-shared component in TIAA can learn the latent cross-modal asynchronous relationships between each modality and fused important tokens, which can alleviate the interference of other tokens from the asynchronous modalities.

More precisely, as shown in Figure \ref{ms_attn}, it is observed that lip and hand modalities exhibit similar modality-shared attention score matrices of the TIAA module, indicating the proposed method can learn the consistent latent relationships for different modalities. For each chunk of lip and hand, modality-shared attentions of lip and hand can focus on the same important tokens with consistent semantic information. Thus, benefiting from the cross-attention based on fused important tokens, our method can well align the semantic relationships for tokens of lip and hand movements, and can capture similar coarse-grained temporal dependencies for different modalities.

\subsection{Ablation Studies}
\label{ablation}
To systematically analyze the effectiveness of each component in the proposed EcoCued, an extensive ablation study is conducted from various perspectives on the Chinese CS dataset under the single-cuer setting. The main experimental results are illustrated in Table \ref{component} and \ref{ablation_std}. The main observations are reported as follows.

\begin{table}[!t]
\caption{Ablation study for different components of EcoCued. The chunk size is 32 and $k$ is 8. TIAA is combining modality-specific and modality-shared attentions.}
\begin{center}
\begin{tabular}{l|c|c|c}
\hline
Components & $k$ & CER   & WER     \\ \hline
Modality-specific      & 8 &  24.4  & 61.3  \\ \hline
Modality-shared        & 8 &  27.1   &  67.6  \\ \hline
Modality-specific + ConAgg     & 8 &  21.1  & 53.9  \\ \hline
Modality-shared + ConAgg       & 8 &  21.2   & 55.7 \\ \hline
TIAA                   & 8 &  15.6  & 40.2   \\ \hline
TIAA + Fusion          & 8 &  14.2  & 39.5  \\ \hline
TIAA + ConAgg         & 8 &  14.5  & 40.7  \\ \hline
TIAA + Fusion + ConAgg     & 8 &  10.0  & 29.7   \\ \hline
\end{tabular}
\end{center}
\label{component}
\end{table}

% \begin{table}[!t]
% \caption{Ablation study for different chunk sizes. The complexity would be increased when the chunk size is too large or too small. The chunk size can refer to the video FPS to allow a chunk to cover a full hand movement.}
% \begin{center}
% \begin{tabular}{c|c|c|c}
% \hline
% Chunk Size & $k$ & CER   & WER \\ \hline

% \end{tabular}
% \end{center}
% \label{chunk}
% \end{table}

% (1) Modality-specific attention contains fine-grained information, but is limited in the local chunk. Modality-shared attention contains the full dependency over the sequence but ignores the fine-grained relations among tokens. Thus, the performance improvement provided by modality-specific or modality-shared attention is limited. 

\textbf{Impact of Different Components.} For the ablation studies on the Chinese CS dataset in Table \ref{component}, it is observed that both modality-specific and modality-shared branches are necessary for better recognition performance. If either modality-specific or modality-shared branch is not utilized, the performance drop is about 10\% on CER and 30\% on WER. When combining these two attention branches, further performance improvement can be obtained on both CER and WER evaluations. Typically, the adoption of the ConAgg module provides an additional reduction on CER and WER evaluations. The main reason covers the following three points: (1)  Modality-specific attention contains fine-grained information, but is limited in the local chunk. Modality-shared attention can further capture information across different chunks of both modalities. (2) The performance can be improved by combining modality-specific/shared and ConAgg for enhancing the spatial relations, but still suffers from insufficient information due to lack of the multi-modal interaction. (3) TIAA-based multi-modal fusion can further decrease the error rate. (4) ConAgg not only enhances the interaction of spatial information, but also makes a better exploitation for both modality-specific and modality-shared information. Thus, combining all of them can achieve the best performance.

\begin{table}[!t]
\caption{Ablation study for different chunk sizes, different choices of $k$, gated hidden projection, and multi-modal fusion.}
\begin{center}
\begin{tabular}{c|c|c|c|c|c}
\hline
Chunk size & $k$   & Gate  & Fusion     & CER   & WER     \\ \hline
16     & 8 & \Checkmark    & \Checkmark &  11.5  & 33.7     \\ \hline
32     & 8 & \Checkmark    & \Checkmark &  10.0  & 29.7     \\ \hline
32     & 8  & \XSolidBrush  & \Checkmark      &  43.4  & 79.9  \\ \hline
32     & 8  & \Checkmark    & \Checkmark      &  10.0  & 29.7      \\ \hline
32     & 8  & \Checkmark    & \XSolidBrush    &  14.5  & 40.7  \\ \hline
48     & 8 & \Checkmark    & \Checkmark &  11.2  & 32.1     \\ \hline
64     & 8 & \Checkmark    & \Checkmark &  10.2  &  29.9    \\ \hline
96     & 4  & \Checkmark    & \Checkmark      &  10.8  & 31.8      \\ \hline
96     & 8  & \Checkmark    & \Checkmark      &  9.3   & 27.1  \\ \hline
96     & 16 & \Checkmark    & \Checkmark      &  10.1  & 29.3   \\ \hline
96     & 32 & \Checkmark    & \Checkmark      &  10.0  &  28.9   \\ \hline
96     & 48 & \Checkmark    & \Checkmark      &  10.4  &  29.9   \\ \hline
\end{tabular}
\end{center}
\label{ablation_std}
\end{table}

\begin{table}[!t]
    \caption{Performance comparison on Chinese CS dataset using the single modality.}
    \begin{center}
    \begin{tabular}{l|c|c|c|c|c}
    \hline
    \multicolumn{2}{c|}{Method} & \multicolumn{4}{c}{Chinese} \\ \hline
    \multicolumn{2}{c|}{\#Modality}  & \multicolumn{2}{c|}{only lip} & \multicolumn{2}{c}{only hand}  \\ \hline
    Metrics           & Param(M)& CER   & WER   & CER   & WER        \\ \hline
    CNN + FLASH       & 7.7     & 64.9  & 98.0  & 62.6  & 99.0       \\ \hline
    CNN + Linformer   & 8.9     & 64.4  & 97.1  & 64.3  & 99.7       \\ \hline
    CNN + Performer   & 11.0    & 64.1  & 99.3  & 55.4  & 95.0       \\ \hline
    CNN + Cosformer   & 7.2     & 68.9  & 97.5  & 62.2  & 94.5       \\ \hline
    Ours              & \textbf{6.6}  & \textbf{41.4}   & \textbf{77.7}  &  \textbf{29.8}   &  \textbf{62.1}      \\ \hline
    \end{tabular}
    \end{center}
    \label{single}
\end{table}
    
\textbf{Impact of Chunk Size.} The chunk size influences both the performance quality and the complexity of EcoCued. As shown in Table \ref{ablation_std}, it is observed larger chunk sizes can perform better with fixed $k$ but lead to higher complexity. When using larger chunk sizes, the computation complexity of modality-specific attention would be increased, while the complexity of modality-shared attention is reduced due to the decreased chunk number. In the case where chunk size is equal to one, the modality-shared attention degenerates as the quadratic self-attention. In the case where chunk size is equal to the sequence length, the modality-specific attention becomes the quadratic self-attention. Both of these cases suffer from inefficient training. Thus, the choice of chunk size would affect the trade-off between modality-specific and modality-shared attentions. Overall, the computational complexity of TIAA would be increased if the chunk size is too large or too small. To preserve important motion information, a chunk should be medium to contain a full hand movement from the previous shape to the next one, referring to the video FPS.

\textbf{Impact of Top-$k$ Selection.} The top-$k$ selection can influence modeling global dependency and complexity. When using larger $k$, the computation complexity of modality-shared attention would be increased. In the case where $k$ is equal to the chunk size, the modality-shared attention of EcoCued degenerates as the quadratic self-attention. As shown in Table \ref{ablation_std}, it is observed the performance is not sensitive to the choices of $k$. The main reason may lie in that there exist significant redundancies within each local chunk due to minor motion changes between consecutive frames. Besides, if $k$ is too small or too large, the performance would be decreased due to dropping information or redundant noisy information.

\textbf{Impact of Gated Hidden Projection.} 
The self-attention mechanism is sensitive to the over-fitting risk and requires much training data to alleviate this problem, while the data scale of the CS dataset is relatively smaller. Gate hidden projection can restrict the information flow from the input to the output projection for self-attention modeling, which is an efficient regularization technique. Here we study the importance of using a gate mechanism in EcoCued. To achieve this, we replace the gate hidden projection using vanilla linear projection, which has the same linear complexity. Table \ref{ablation_std} shows the performance comparison with and without the gate mechanism. It is observed that there is a significant performance drop without a gate mechanism, confirming the importance of gate hidden projection in our EcoCued method. 

\textbf{Impact of Multi-modal Fusion.} 
In this part, we study the effectiveness of multi-modal fusion in EcoCued. To this end, we remove the multi-modal fusion, \textit{i.e.}, each modality utilizes inter-modality information without multi-modal fusion, ignoring the information flow between different modalities. Table \ref{ablation_std} reports the performance comparison with and without multi-modal interaction. It shows that the EcoCued can benefit from the cross-modal interaction, indicating the effectiveness of cross-modal interaction. One advantage is that the multi-modal interaction is based on the important tokens of each modality, while does not introduce additional parameters.

\textbf{Impact of Different Modalities.} In this part, we study the effectiveness of different modalities for the ACSR task. As shown in Table \ref{single} under the single-cuer setting. The results indicate that our method can still outperform the comparison methods on the single modality. Besides, we observe that there exists significant performance drop using single modality for training, where hand modality exhibits higher recognition accuracy than lip modality.
\section{Acknowledgment}
This work is supported by the National Natural Science Foundation of China (No. 62101351) and Tencent AI Lab Rhino Bird Funding (RBFR2023014).

\section{Conclusion}
In this work, we propose a computation and parameter-efficient multi-modal fusion method called EcoCued for the ACSR task. Specially, we present a novel Token-Importance-Aware Attention mechanism (TIAA) with a novel token utilization rate (TUR) to select the important tokens from the multi-modal feature streams. To capture long-range dependency, TIAA decomposes full attention into the modality-specific and modality-shared contextual information for a higher-quality self-attention mechanism. Then it conducts the efficient cross-modal interaction for the modality-shared component over the important tokens of different modalities. Furthermore, a Convolution-based Aggregation (ConAgg) is presented to capture the spatial relation for the TIAA mechanism. Finally, a light-weight gated hidden projection is designed to control the feature flow through the TIAA module. The proposed method achieves SOTA performance on the Mandarin Chinese, French, and British CS benchmarks, compared with existing transformer-based methods and previous ACSR methods. Importantly, our method can reduce the computational complexity of the self-attention from $\mathcal{O}(T^2)$ to $\mathcal{O}(T)$ with a light-weight architecture. By ablation studies, multi-modal fusion can be efficiently achieved by focusing on the important features of each modality for ACSR task due to low-rank property, where spatial interaction can further enhance the information for fused modalities. In the future, we will explore large-scale multi-modal pre-training methods for the ACSR task.

\bibliographystyle{IEEEtran}
\bibliography{reference}

\begin{thebibliography}{10}
\providecommand{\url}[1]{#1}
\csname url@samestyle\endcsname
\providecommand{\newblock}{\relax}
\providecommand{\bibinfo}[2]{#2}
\providecommand{\BIBentrySTDinterwordspacing}{\spaceskip=0pt\relax}
\providecommand{\BIBentryALTinterwordstretchfactor}{4}
\providecommand{\BIBentryALTinterwordspacing}{\spaceskip=\fontdimen2\font plus
\BIBentryALTinterwordstretchfactor\fontdimen3\font minus \fontdimen4\font\relax}
\providecommand{\BIBforeignlanguage}[2]{{%
\expandafter\ifx\csname l@#1\endcsname\relax
\typeout{** WARNING: IEEEtran.bst: No hyphenation pattern has been}%
\typeout{** loaded for the language `#1'. Using the pattern for}%
\typeout{** the default language instead.}%
\else
\language=\csname l@#1\endcsname
\fi
#2}}
\providecommand{\BIBdecl}{\relax}
\BIBdecl

\bibitem{cornett1967cued}
R.~O. Cornett, ``Cued speech,'' \emph{American Annals of the Deaf}, pp. 3--13, 1967.

\bibitem{liu2019pilot}
L.~Liu and G.~Feng, ``A pilot study on mandarin chinese cued speech,'' \emph{American Annals of the Deaf}, vol. 164, no.~4, pp. 496--518, 2019.

\bibitem{liu2018automatic}
L.~Liu, G.~Feng, and D.~Beautemps, ``Automatic temporal segmentation of hand movements for hand positions recognition in french cued speech,'' in \emph{IEEE International Conference on Acoustics, Speech and Signal Processing (ICASSP)}.\hskip 1em plus 0.5em minus 0.4em\relax IEEE, 2018, pp. 3061--3065.

\bibitem{liu2019automatic}
L.~Liu, J.~Li, G.~Feng, and X.-P.~S. Zhang, ``Automatic detection of the temporal segmentation of hand movements in british english cued speech.'' in \emph{INTERSPEECH}, 2019, p. 2285–2289.

\bibitem{zhang2023cuing}
Y.~Zhang, L.~Liu, and L.~Liu, ``Cuing without sharing: A federated cued speech recognition framework via mutual knowledge distillation,'' in \emph{Proceedings of the 31st ACM International Conference on Multimedia}, 2023, p. 8781–8789.

\bibitem{heracleous2012continuous}
P.~Heracleous, D.~Beautemps, and N.~Hagita, ``Continuous phoneme recognition in cued speech for french,'' in \emph{Proceedings of the European Signal Processing Conference (EUSIPCO)}.\hskip 1em plus 0.5em minus 0.4em\relax IEEE, 2012, pp. 2090--2093.

\bibitem{liu2018visual}
L.~Liu, T.~Hueber, G.~Feng, and D.~Beautemps, ``Visual recognition of continuous cued speech using a tandem cnn-hmm approach.'' in \emph{Interspeech}, 2018, pp. 2643--2647.

\bibitem{wang2021attention}
J.~Wang, N.~Gu, M.~Yu, X.~Li, Q.~Fang, and L.~Liu, ``An attention self-supervised contrastive learning based three-stage model for hand shape feature representation in cued speech,'' in \emph{Proceedings of Interspeech}, 2021, pp. 626--630.

\bibitem{liu2020re}
L.~\vspace{0mm}Liu, G.~Feng, D.~Beautemps, and X.-P. Zhang, ``Re-synchronization using the hand preceding model for multi-modal fusion in automatic continuous cued speech recognition,'' \emph{IEEE Transactions on Multimedia}, vol.~23, pp. 292--305, 2020.

\bibitem{wang2021cross}
J.~Wang, Z.~Tang, X.~Li, M.~Yu, Q.~Fang, and L.~Liu, ``Cross-modal knowledge distillation method for automatic cued speech recognition,'' in \emph{Interspeech}, 2021, p. 2986–2990.

\bibitem{wei2020multi}
X.~Wei, T.~Zhang, Y.~Li, Y.~Zhang, and F.~Wu, ``Multi-modality cross attention network for image and sentence matching,'' in \emph{Proceedings of the IEEE/CVF conference on Computer Vision and Pattern Recognition}, 2020, pp. 10\,941--10\,950.

\bibitem{vaswani2017attention}
A.~Vaswani, N.~Shazeer, N.~Parmar, J.~Uszkoreit, L.~Jones, A.~N. Gomez, {\L}.~Kaiser, and I.~Polosukhin, ``Attention is all you need,'' \emph{Advances in Neural Information Processing Systems}, vol.~30, 2017.

\bibitem{qian2022audio}
X.~Qian, Z.~Wang, J.~Wang, G.~Guan, and H.~Li, ``Audio-visual cross-attention network for robotic speaker tracking,'' \emph{IEEE/ACM Transactions on Audio, Speech, and Language Processing}, vol.~31, pp. 550--562, 2022.

\bibitem{liu2023cross}
L.~Liu and L.~Liu, ``Cross-modal mutual learning for cued speech recognition,'' in \emph{IEEE International Conference on Acoustics, Speech and Signal Processing (ICASSP)}.\hskip 1em plus 0.5em minus 0.4em\relax IEEE, 2023, pp. 1--5.

\bibitem{katharopoulos2020transformers}
A.~Katharopoulos, A.~Vyas, N.~Pappas, and F.~Fleuret, ``Transformers are rnns: Fast autoregressive transformers with linear attention,'' in \emph{International Conference on Machine Learning}.\hskip 1em plus 0.5em minus 0.4em\relax PMLR, 2020, pp. 5156--5165.

\bibitem{mehtadelight}
S.~Mehta, M.~Ghazvininejad, S.~Iyer, L.~Zettlemoyer, and H.~Hajishirzi, ``Delight: Deep and light-weight transformer,'' in \emph{International Conference on Learning Representations}, 2022.

\bibitem{wu2022flowformer}
H.~Wu, J.~Wu, J.~Xu, J.~Wang, and M.~Long, ``Flowformer: Linearizing transformers with conservation flows,'' \emph{arXiv preprint arXiv:2202.06258}, 2022.

\bibitem{peng2020random}
H.~Peng, N.~Pappas, D.~Yogatama, R.~Schwartz, N.~Smith, and L.~Kong, ``Random feature attention,'' in \emph{International Conference on Learning Representations}, 2020.

\bibitem{tay2022efficient}
Y.~Tay, M.~Dehghani, D.~Bahri, and D.~Metzler, ``Efficient transformers: A survey,'' \emph{ACM Computing Surveys}, vol.~55, no.~6, pp. 1--28, 2022.

\bibitem{qin2021cosformer}
Z.~Qin, W.~Sun, H.~Deng, D.~Li, Y.~Wei, B.~Lv, J.~Yan, L.~Kong, and Y.~Zhong, ``cosformer: Rethinking softmax in attention,'' in \emph{International Conference on Learning Representations}, 2021.

\bibitem{choromanski2020rethinking}
K.~M. Choromanski, V.~Likhosherstov, D.~Dohan, X.~Song, A.~Gane, T.~Sarlos, P.~Hawkins, J.~Q. Davis, A.~Mohiuddin, L.~Kaiser \emph{et~al.}, ``Rethinking attention with performers,'' in \emph{International Conference on Learning Representations}, 2020.

\bibitem{wang2020linformer}
S.~Wang, B.~Z. Li, M.~Khabsa, H.~Fang, and H.~Ma, ``Linformer: Self-attention with linear complexity,'' \emph{arXiv preprint arXiv:2006.04768}, 2020.

\bibitem{child2019generating}
R.~Child, S.~Gray, A.~Radford, and I.~Sutskever, ``Generating long sequences with sparse transformers,'' \emph{arXiv preprint arXiv:1904.10509}, 2019.

\bibitem{kitaev2019reformer}
N.~Kitaev, L.~Kaiser, and A.~Levskaya, ``Reformer: The efficient transformer,'' in \emph{International Conference on Learning Representations}, 2019.

\bibitem{zaheer2020big}
M.~Zaheer, G.~Guruganesh, K.~A. Dubey, J.~Ainslie, C.~Alberti, S.~Ontanon, P.~Pham, A.~Ravula, Q.~Wang, L.~Yang \emph{et~al.}, ``Big bird: Transformers for longer sequences,'' \emph{Advances in Neural Information Processing Systems}, vol.~33, pp. 17\,283--17\,297, 2020.

\bibitem{wupay}
F.~Wu, A.~Fan, A.~Baevski, Y.~Dauphin, and M.~Auli, ``Pay less attention with lightweight and dynamic convolutions,'' in \emph{International Conference on Learning Representations}, 2019.

\bibitem{dauphin2017language}
Y.~N. Dauphin, A.~Fan, M.~Auli, and D.~Grangier, ``Language modeling with gated convolutional networks,'' in \emph{International Conference on Machine Learning}.\hskip 1em plus 0.5em minus 0.4em\relax PMLR, 2017, pp. 933--941.

\bibitem{wulite}
Z.~Wu, Z.~Liu, J.~Lin, Y.~Lin, and S.~Han, ``Lite transformer with long-short range attention,'' in \emph{International Conference on Learning Representations}, 2020.

\bibitem{hua2022transformer}
W.~Hua, Z.~Dai, H.~Liu, and Q.~Le, ``Transformer quality in linear time,'' in \emph{International Conference on Machine Learning}.\hskip 1em plus 0.5em minus 0.4em\relax PMLR, 2022, pp. 9099--9117.

\bibitem{papadimitriou2021fully}
K.~Papadimitriou and G.~Potamianos, ``A fully convolutional sequence learning approach for cued speech recognition from videos,'' in \emph{European Signal Processing Conference (EUSIPCO)}.\hskip 1em plus 0.5em minus 0.4em\relax IEEE, 2021, pp. 326--330.

\bibitem{Radosavovic_2020_CVPR}
I.~Radosavovic, R.~P. Kosaraju, R.~Girshick, K.~He, and P.~Doll{\'a}r, ``Designing network design spaces,'' in \emph{Proceedings of the IEEE/CVF Conference on Computer Vision and Pattern Recognition (CVPR)}, June 2020.

\bibitem{siriwardhana2020jointly}
S.~Siriwardhana, A.~Reis, R.~Weerasekera, and S.~Nanayakkara, ``Jointly fine-tuning" bert-like" self supervised models to improve multimodal speech emotion recognition,'' in \emph{Interspeech}, 2020.

\bibitem{priyasad2020attention}
D.~Priyasad, T.~Fernando, S.~Denman, S.~Sridharan, and C.~Fookes, ``Attention driven fusion for multi-modal emotion recognition,'' in \emph{ICASSP}, 2020, pp. 3227--3231.

\bibitem{you2020towards}
C.~You, N.~Chen, F.~Liu, D.~Yang, and Y.~Zou, ``Towards data distillation for end-to-end spoken conversational question answering,'' \emph{arXiv preprint arXiv:2010.08923}, 2020.

\bibitem{you2021knowledge}
C.~You, N.~Chen, and Y.~Zou, ``Knowledge distillation for improved accuracy in spoken question answering,'' in \emph{IEEE International Conference on Acoustics, Speech and Signal Processing (ICASSP)}, 2021, pp. 7793--7797.

\bibitem{you2022end}
C.~You, N.~Chen, F.~Liu, S.~Ge, X.~Wu, and Y.~Zou, ``End-to-end spoken conversational question answering: Task, dataset and model,'' in \emph{Findings of the Association for Computational Linguistics: NAACL 2022}, 2022, pp. 1219--1232.

\bibitem{you2020contextualized}
C.~You, N.~Chen, and Y.~Zou, ``Contextualized attention-based knowledge transfer for spoken conversational question answering,'' \emph{arXiv preprint arXiv:2010.11066}, 2020.

\bibitem{chen2021self}
N.~Chen, C.~You, and Y.~Zou, ``Self-supervised dialogue learning for spoken conversational question answering,'' \emph{arXiv preprint arXiv:2106.02182}, 2021.

\bibitem{you2021self}
C.~You, N.~Chen, and Y.~Zou, ``Self-supervised contrastive cross-modality representation learning for spoken question answering,'' in \emph{Findings of the Association for Computational Linguistics: EMNLP 2021}, 2021, pp. 28--39.

\bibitem{you2021mrd}
{You, Chenyu and Chen, Nuo and Zou, Yuexian}, ``Mrd-net: Multi-modal residual knowledge distillation for spoken question answering,'' in \emph{IJCAI}, 2021, pp. 3985--3991.

\bibitem{burger2005cued}
T.~Burger, A.~Caplier, and S.~Mancini, ``Cued speech hand gestures recognition tool,'' in \emph{2005 13th European Signal Processing Conference}.\hskip 1em plus 0.5em minus 0.4em\relax IEEE, 2005, pp. 1--4.

\bibitem{stillittano2013lip}
S.~Stillittano, V.~Girondel, and A.~Caplier, ``Lip contour segmentation and tracking compliant with lip-reading application constraints,'' \emph{Machine Vision and Applications}, vol.~24, no.~1, pp. 1--18, 2013.

\bibitem{heracleous2010cued}
P.~Heracleous, D.~Beautemps, and N.~Aboutabit, ``Cued speech automatic recognition in normal-hearing and deaf subjects,'' \emph{Speech Communication}, vol.~52, no.~6, pp. 504--512, 2010.

\bibitem{liu2019novel}
L.~Liu, G.~Feng, D.~Beautemps, and X.-P. Zhang, ``A novel resynchronization procedure for hand-lips fusion applied to continuous french cued speech recognition,'' in \emph{Proceedings of the European Signal Processing Conference (EUSIPCO)}, 2019, pp. 1--5.

\bibitem{dlib}
\BIBentryALTinterwordspacing
{dlib}. [Online]. Available: \url{http://dlib.net}
\BIBentrySTDinterwordspacing

\bibitem{mediapipe}
\BIBentryALTinterwordspacing
{mediapipe}. [Online]. Available: \url{https://mediap}
\BIBentrySTDinterwordspacing

\bibitem{xu2023multimodal}
P.~Xu, X.~Zhu, and D.~A. Clifton, ``Multimodal learning with transformers: A survey,'' \emph{IEEE Transactions on Pattern Analysis and Machine Intelligence}, 2023.

\bibitem{qiu2019blockwise}
J.~Qiu, H.~Ma, O.~Levy, S.~W.-t. Yih, S.~Wang, and J.~Tang, ``Blockwise self-attention for long document understanding,'' in \emph{International Conference on Learning Representations}, 2019.

\bibitem{vyas2020fast}
A.~Vyas, A.~Katharopoulos, and F.~Fleuret, ``Fast transformers with clustered attention,'' \emph{Advances in Neural Information Processing Systems}, vol.~33, pp. 21\,665--21\,674, 2020.

\bibitem{rahimi2007random}
A.~Rahimi and B.~Recht, ``Random features for large-scale kernel machines,'' \emph{Advances in Neural Information Processing Systems}, vol.~20, 2007.

\bibitem{williams2000using}
C.~Williams and M.~Seeger, ``Using the nystr{\"o}m method to speed up kernel machines,'' \emph{Advances in Neural Information Processing Systems}, vol.~13, 2000.

\bibitem{lu2021soft}
J.~Lu, J.~Yao, J.~Zhang, X.~Zhu, H.~Xu, W.~Gao, C.~Xu, T.~Xiang, and L.~Zhang, ``Soft: softmax-free transformer with linear complexity,'' \emph{Advances in Neural Information Processing Systems}, vol.~34, pp. 21\,297--21\,309, 2021.

\bibitem{zeng2021you}
Z.~Zeng, Y.~Xiong, S.~Ravi, S.~Acharya, G.~M. Fung, and V.~Singh, ``You only sample (almost) once: Linear cost self-attention via bernoulli sampling,'' in \emph{International Conference on Machine Learning}.\hskip 1em plus 0.5em minus 0.4em\relax PMLR, 2021, pp. 12\,321--12\,332.

\bibitem{chen2021crossvit}
C.-F.~R. Chen, Q.~Fan, and R.~Panda, ``Crossvit: Cross-attention multi-scale vision transformer for image classification,'' in \emph{Proceedings of the IEEE/CVF International Conference on Computer Vision}, 2021, pp. 357--366.

\bibitem{ramachandran2017searching}
P.~Ramachandran, B.~Zoph, and Q.~V. Le, ``Searching for activation functions,'' \emph{arXiv preprint arXiv:1710.05941}, 2017.

\bibitem{he2016deep}
K.~He, X.~Zhang, S.~Ren, and J.~Sun, ``Deep residual learning for image recognition,'' in \emph{Proceedings of the IEEE Conference on Computer Vision and Pattern Recognition}, 2016, pp. 770--778.

\bibitem{gao2023novel}
L.~Gao, S.~Huang, and L.~Liu, ``A novel interpretable and generalizable re-synchronization model for cued speech based on a multi-cuer corpus,'' \emph{arXiv preprint arXiv:2306.02596}, 2023.

\bibitem{liu2022objective}
L.~Liu, G.~Feng, X.~Ren, and X.~Ma, ``Objective hand complexity comparison between two mandarin chinese cued speech systems,'' in \emph{2022 13th International Symposium on Chinese Spoken Language Processing (ISCSLP)}.\hskip 1em plus 0.5em minus 0.4em\relax IEEE, 2022, p. 215–219.

\bibitem{sankar2022multistream}
S.~Sankar, D.~Beautemps, and T.~Hueber, ``Multistream neural architectures for cued speech recognition using a pre-trained visual feature extractor and constrained ctc decoding,'' in \emph{IEEE International Conference on Acoustics, Speech and Signal Processing (ICASSP)}.\hskip 1em plus 0.5em minus 0.4em\relax IEEE, 2022, pp. 8477--8481.

\bibitem{cubuk2020randaugment}
E.~D. Cubuk, B.~Zoph, J.~Shlens, and Q.~V. Le, ``Randaugment: Practical automated data augmentation with a reduced search space,'' in \emph{Proceedings of the IEEE/CVF Conference on Computer Vision and Pattern Recognition Workshops}, 2020, pp. 702--703.

\bibitem{afouras2018deep}
T.~Afouras, J.~S. Chung, A.~Senior, O.~Vinyals, and A.~Zisserman, ``Deep audio-visual speech recognition,'' \emph{IEEE Transactions on Pattern Analysis and Machine Intelligence}, vol.~44, no.~12, pp. 8717--8727, 2018.

\bibitem{yu2020audio}
J.~Yu, S.-X. Zhang, J.~Wu, S.~Ghorbani, B.~Wu, S.~Kang, S.~Liu, X.~Liu, H.~Meng, and D.~Yu, ``Audio-visual recognition of overlapped speech for the lrs2 dataset,'' in \emph{IEEE International Conference on Acoustics, Speech and Signal Processing (ICASSP)}, 2020, pp. 6984--6988.

\bibitem{ma2021end}
P.~Ma, S.~Petridis, and M.~Pantic, ``End-to-end audio-visual speech recognition with conformers,'' in \emph{IEEE International Conference on Acoustics, Speech and Signal Processing (ICASSP)}, 2021, pp. 7613--7617.

\bibitem{attina2004pilot}
V.~Attina, D.~Beautemps, M.-A. Cathiard, and M.~Odisio, ``A pilot study of temporal organization in cued speech production of french syllables: rules for a cued speech synthesizer,'' \emph{Speech Communication}, vol.~44, no.~1, pp. 197--214, 2004.

\bibitem{attina2005temporal}
V.~Attina, M.-A. Cathiard, and D.~Beautemps, ``Temporal measures of hand and speech coordination during french cued speech production,'' in \emph{International Gesture Workshop}.\hskip 1em plus 0.5em minus 0.4em\relax Springer, 2005, pp. 13--24.

\end{thebibliography}
\end{document}